\documentclass[10pt,twocolumn,letterpaper]{article}

\usepackage{cvpr}
\usepackage{times}
\usepackage{epsfig}
\usepackage{graphicx}
\usepackage{amsmath}
\usepackage{amssymb}
\usepackage{graphicx}  \graphicspath{{images/}}
\usepackage{algpseudocode}
\usepackage[normalem]{ulem}
\usepackage{makecell}
\usepackage{rotating}
\usepackage{booktabs}
\usepackage{gensymb}
\usepackage[dvipsnames,svgnames,x11names]{xcolor}

\usepackage[utf8]{inputenc} 
\usepackage[T1]{fontenc}    
\usepackage{url}            
\usepackage{booktabs}       
\usepackage{amsfonts}       
\usepackage{amsmath}
\usepackage{nicefrac}       
\usepackage{microtype}      
\usepackage{graphicx}
\usepackage{color}
\usepackage{algorithm2e}
\usepackage{subfig}
\usepackage{makecell}
\usepackage{appendix}
\usepackage{amsthm}
\newtheorem{theorem}{Theorem}
\newtheorem{proposition}{Proposition}

\usepackage[pagebackref=true,breaklinks=true,letterpaper=true,colorlinks,bookmarks=false]{hyperref}

\cvprfinalcopy 

\def\cvprPaperID{092} 

\usepackage[dvipsnames,svgnames,x11names]{xcolor}
\newcolumntype{C}[1]{>{\centering\let\newline\\\arraybackslash\hspace{0pt}}m{#1}}
\newcommand{\setmode}[1]{\def\mode{#1}}
\setmode{draft} 
\long\def\IGNORE#1{} \long\def\COMMENT#1{}
\def\authornote#1#2#3{{\textcolor{#2}{\textsl{\small#1:[*#3*]}}}}
\ifthenelse{\equal{\mode}{draft}} { 
	\newcommand{\arnote}[1]{\authornote{AR}{Red}{#1}} 
	\newcommand{\mbnote}[1]{\authornote{MB}{Cyan}{#1}} 
    \newcommand{\vjnote}[1]{\authornote{VJ}{Orange}{#1}} 
	\newcommand{\dsnote}[1]{\authornote{DS}{Emerald}{#1}} 
	\newcommand{\jwnote}[1]{\authornote{JW}{Green}{#1}} 
    \newcommand{\khnote}[1]{\authornote{KH}{Blue}{#1}} 
} {}

\ifthenelse{\equal{\mode}{final}} {
	\newcommand{\arnote}[1]{}	
	\newcommand{\vjnote}[1]{}
    \newcommand{\dsnote}[1]{}
	\newcommand{\khnote}[1]{}
	\newcommand{\jwnote}[1]{}
	\newcommand{\khnote}[1]{}
	\newcommand{\mbnote}[1]{}
	\typeout{************* MODE: Final}
} {}

\newcommand\blfootnote[1]{%
  \begingroup
  \renewcommand\thefootnote{}\footnote{#1}%
  \addtocounter{footnote}{-1}%
  \endgroup
}
\begin{document}

\title{Competitive Collaboration: Joint Unsupervised Learning of Depth, Camera Motion, Optical Flow and Motion Segmentation }


%

\author{
 Anurag Ranjan$^1$ \quad
  Varun Jampani$^2$ \quad
  Lukas Balles$^1$ \\
  Kihwan Kim $^2$ \quad
  Deqing Sun $^2$ \quad
  Jonas Wulff $^{1,3}$ \quad
  Michael J. Black$^1$ \\
    $^1$Max Planck Institute for Intelligent Systems \quad
    $^2$NVIDIA \quad
    $^3$MIT\\
    \texttt{\{aranjan, lballes, jwulff, black\}@tuebingen.mpg.de} \\
        \texttt{\{vjampani, kihwank, deqings\}@nvidia.com}\\
}

\maketitle

\begin{abstract}
We address the unsupervised learning of several interconnected problems in low-level vision:  single view depth prediction, camera motion estimation, optical flow, and segmentation of a video into the static scene and moving regions.
Our key insight is that these four fundamental vision problems are coupled through geometric constraints.
Consequently, learning to solve them together simplifies the problem because the solutions can reinforce each other.
We go beyond previous work by exploiting geometry more explicitly and segmenting the scene into static and moving regions.
To that end, we introduce {\em Competitive Collaboration}, a framework that facilitates the coordinated training of multiple specialized neural networks to solve complex problems.
Competitive Collaboration works much like expectation-maximization, but with neural networks that act as both competitors to explain pixels that correspond to static or moving regions, and as collaborators through a moderator that assigns pixels to be either static or independently moving.
Our novel method integrates all these problems in a common framework and simultaneously reasons about the segmentation of the scene into moving objects and the static background, the camera motion, depth of the static scene structure, and the optical flow of moving objects.
Our model is trained without any supervision and achieves state-of-the-art performance among joint unsupervised methods on all sub-problems.

\end{abstract}
\blfootnote{This project was formerly referred by \textit{Adversarial Collaboration: Joint Unsupervised Learning of Depth, Camera Motion, Optical Flow and Motion Segmentation}}.
\section{Introduction}
\label{sec:intro}

Deep learning methods have achieved  state-of-the-art results on computer vision problems with supervision using large amounts of data \cite{he2017maskrcnn,krizhevsky2012imagenet,long2015fully}.
However, for many vision problems requiring dense, continuous-valued outputs, it is either impractical or expensive to gather ground truth data \cite{Geiger2012CVPR}. 
We consider four such problems in this paper: single view depth prediction, camera motion estimation, optical flow, and motion segmentation.
Previous work has approached these problems with supervision using real \cite{eigen2014depth} and synthetic data \cite{dosovitskiy2015flownet}.
However there is always a realism gap between synthetic and real data, and real data is limited or inaccurate.
For example, depth ground truth obtained using LIDAR \cite{Geiger2012CVPR} is sparse.
Furthermore, there are no sensors that provide ground truth optical flow, so all existing datasets with real imagery are limited or approximate \cite{Baker:IJCV:11,Geiger2012CVPR,Janai2017CVPR}.
Motion segmentation ground truth currently requires manual
labeling of all pixels in an image \cite{Perazzi2016davis}.

\begin{figure}[t]
  \centering
\includegraphics[width=\linewidth]{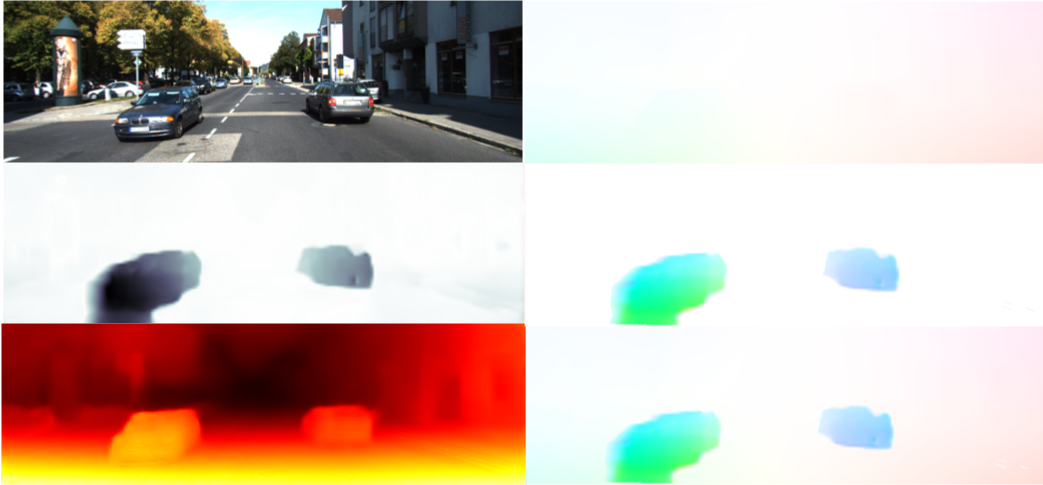} \\
	\caption{\textbf{Unsupervised Learning of Depth, Camera Motion, Optical Flow and Motion Segmentation.} Left, top to bottom: sample image, soft masks representing motion segmentation, estimated depth map.
Right, top to bottom: static scene optical flow, segmented flow in the moving regions and combined optical flow. }
    \vspace{-0.4cm}
    \label{fig:teaser_topright}
\end{figure}

{\bf Problem.}
Recent work has tried to address the problem of limited training data using unsupervised learning \cite{jason2016back2basics, meister2017unflow}.
To learn a mapping from pixels to flow, depth, and camera motion without ground truth is challenging because each of these problems is highly ambiguous.
To address this, additional constraints are needed and the geometric relations between static scenes, camera motion, and optical flow can be exploited.
For example, unsupervised learning of depth and camera motion has been coupled in \cite{zhou2017unsupervised,mahjourian2018googleicp}. They use an explainability mask to exclude evidence that cannot be explained by the static scene assumption.
Yin et al.~\cite{yin2018geonet} extend this to estimate optical flow as well and use forward-backward consistency to reason about unexplained pixels. These methods perform poorly on depth~\cite{zhou2017unsupervised} and optical flow \cite{yin2018geonet} benchmarks.
A key reason is that the constraints applied here do not distinguish or segment objects that move independently like people and cars.
More generally,
not all the data in the unlabeled training set will conform to the model assumptions, and some of it might corrupt the network training.
For instance, the training data for depth and camera motion should not contain independently moving objects.
Similarly, for optical flow, the data should not contain occlusions, which disrupt the commonly used photometric loss.

\begin{figure*}[t]
  \centering
	\includegraphics[width=.8\linewidth]{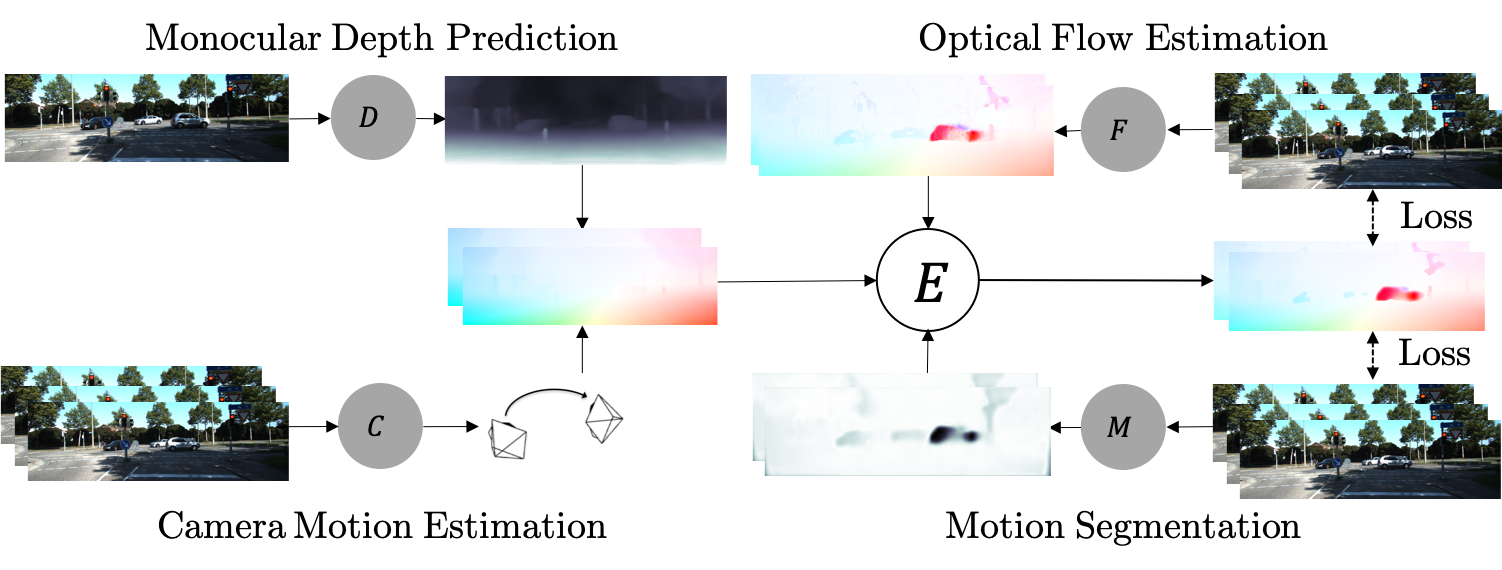}
	\caption{The network $R=(D,C)$ reasons about the scene by estimating optical flow over static regions using depth, $D$, and camera motion, $C$. The optical flow network $F$ estimates flow over the whole image. The motion segmentation network, $M$, masks out static scene pixels from $F$ to produce composite optical flow over the full image. A loss, $E$, using the composite flow is applied over neighboring frames to train all these models jointly. }
    \label{fig:teaser}
    \vspace{-0.3cm}
\end{figure*}

{\bf Idea.}
A typical real-world scene consists of static regions, which do not move in the physical world, and moving objects~\cite{Wulff17cvpr}.
Given depth and camera-motion, we can reason about the static scene in a video sequence. Optical flow, in contrast, reasons about all parts of the scene.
Motion segmentation classifies a scene into static and moving regions.
Our key insight is that these problems are coupled by the geometry and motion of the scene; therefore solving them jointly is synergistic.
We show that by learning jointly from unlabeled data, our coupled networks can partition the dataset and use only the relevant data, resulting in more accurate results than learning without this synergy.

{\bf Approach.}
To address the problem of joint unsupervised learning, we introduce \emph{Competitive Collaboration (CC)}, a generic framework in which networks learn to collaborate and compete, thereby achieving specific goals.
In our specific scenario, Competitive Collaboration is a three player game consisting of two players competing for a resource that is regulated by a third player, moderator.
As shown in Figure~\ref{fig:teaser}, we introduce two players in our framework, the static scene reconstructor, $R=(D,C)$, that reasons about the static scene pixels using depth, $D$, and camera motion, $C$; and a moving region reconstructor, $F$, that reasons about pixels in the independently moving regions. These two players compete for training data by reasoning about static-scene and moving-region pixels in an image sequence. The competition is moderated by a motion segmentation network, $M$, that segments the static scene and moving regions, and distributes training data to the players. However, the moderator also needs training to ensure a fair competition. Therefore, the players, $R$ and $F$, collaborate to train the moderator, $M$, such that it classifies static and moving regions correctly in alternating phases of the training cycle.
This general framework is similar in spirit to expectation-maximization (EM) but is formulated for neural network training.

{\bf Contributions.} In summary our contributions are:
1) We introduce \emph{Competitive Collaboration}, an unsupervised learning framework where networks act as competitors and collaborators to reach specific goals.
2) We show that jointly training networks with this framework has a synergistic effect on their performance.
3) To our knowledge, our method is the first to use low level information like depth, camera motion and optical flow to solve a segmentation task without any supervision.
4) We achieve state-of-the-art performance on single view depth prediction and camera motion estimation among unsupervised methods. We achieve state of art performance on optical flow among unsupervised methods methods that reason about the geometry of the scene, and introduce the first baseline for fully unsupervised motion segmentation. We even outperform competing methods that use much larger networks \cite{yin2018geonet} and multiple refinement steps such as network cascading \cite{meister2017unflow}.
5) We analyze the convergence properties of our method and give an intuition of its generalization using mixed domain learning on MNIST~\cite{Lecun:1998:MNist} and SVHN~\cite{Netzer:2011:SVHN} digits.
All our models and code are available at \url{https://github.com/anuragranj/cc}.

\section{Related Work}

Our method is a three-player game, consisting of two competitors and a moderator, where the moderator takes the role of a critic and two competitors collaborate to train the moderator. The idea of collaboration can also be seen as neural expectation maximization \cite{greff2017neuralEM} where one model is trained to distribute data to other models.
For unsupervised learning, these ideas have been mainly used to model the data distribution \cite{greff2017neuralEM} and have not been applied to unsupervised training of regression or classification problems.

There is significant recent work on supervised training of single image depth prediction \cite{eigen2014depth}, camera motion estimation \cite{Kendall15iccv} and optical flow estimation \cite{dosovitskiy2015flownet}.
However, as labeling large datasets for continuous-valued regression tasks is not trivial, and the methods often rely on synthetic data~\cite{dosovitskiy2015flownet,mayer2016large}. 
Unsupervised methods have tried to independently solve for optical flow \cite{jason2016back2basics,meister2017unflow,Wulff:GCPR:2018} by minimizing a photometric loss. 
This is highly underconstrained and thus the methods perform poorly.

More recent works ~\cite{mahjourian2018googleicp,Ummenhofer17cvpr,Vijayanarasimhan17arXiv, yin2018geonet, zhou2017unsupervised} have approached estimation of these problems by coupling two or more problems together in an unsupervised learning framework.
Zhou et al.~\cite{zhou2017unsupervised} introduce joint unsupervised learning of ego-motion and depth from multiple unlabeled frames.
To account for moving objects, they learn an explainability mask. However, these masks also capture model failures such as occlusions at depth discontinuities, and are hence not useful for motion segmentation.
Mahjourian et al.~\cite{mahjourian2018googleicp} use a more explicit geometric loss 
to jointly learn depth and camera motion for rigid scenes. Yin et al.~\cite{yin2018geonet} add a refinement network to \cite{zhou2017unsupervised} to also estimate residual optical flow. The estimation of residual flow is designed to account for moving regions, but there is no coupling of the optical flow network with the depth and camera motion networks. Residual optical flow is obtained using a cascaded refinement network, thus preventing other networks from using flow information to improve themselves. Therefore, recent works show good performance either on depth and camera motion \cite{mahjourian2018googleicp,yin2018geonet,zhou2017unsupervised} or on optical flow~\cite{meister2017unflow}, but not on both.
Zou et al.~\cite{zou2018df} exploit consistency between depth and optical flow to improve performance.
The key missing piece that we add is to jointly learn the segmentation of the scene into static and independently-moving regions.
This allows the networks to use geometric constraints where they apply and generic flow where they do not.
Our work introduces a framework where motion segmentation, flow, depth and camera motion models can be coupled and solved jointly to reason about the complete geometric structure and motion of the scene.


Competitive Collaboration can be generalized to problems in which the models have intersecting goals where they can compete and collaborate.
For example, modeling multi-modal distributions can be accomplished using our framework, whereby each competitor learns the distribution over a mode. In fact, the use of expectation-maximization (EM) in computer vision began with the optical flow problem and was used to segment the scene into ``layers'' \cite{Black:IEEE:1993} and was then widely applied to other vision problems.

\section{Competitive Collaboration}
\begin{figure}[t]
  \centering
	\includegraphics[width=\linewidth]{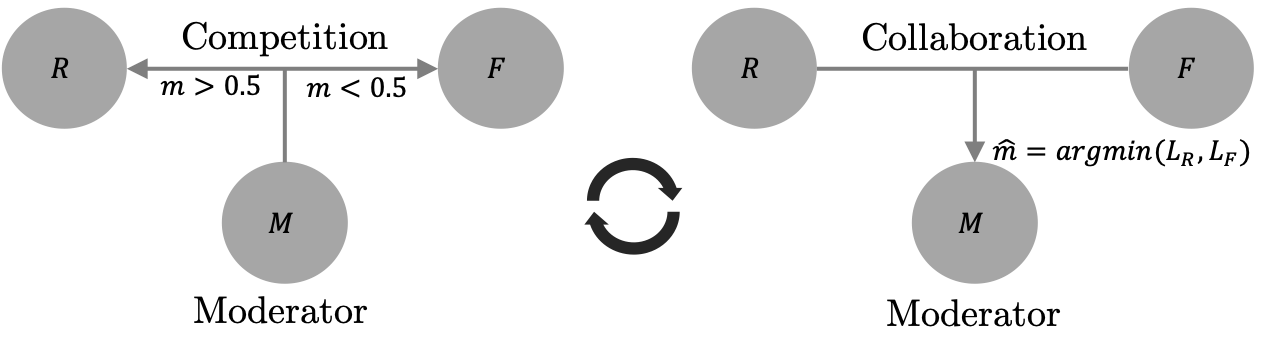}
	\caption{Training cycle of Competitive Collaboration: The moderator $M$ drives two competitors $\{R,F\}$ (first phase, left). Later, the competitors collaborate to train the moderator to ensure fair competition in the next iteration (second phase, right).}
    \label{fig:mac}
\end{figure}

In our context, Competitive Collaboration is formulated as a three-player game consisting of two players competing for a resource that is regulated by a moderator as illustrated in Figure \ref{fig:mac}. Consider an unlabeled training dataset $\mathcal{D} = \{\mathcal{D}_i : i \in \mathbb{N} \}$, which can be partitioned into two disjoint sets. Two players $\{R, F\}$ compete to obtain this data as a resource, and each player tries to partition $\mathcal{D}$ to minimize its loss.
The partition is regulated by the moderator's output $m = M(\mathcal{D}_i), m \in [0,1]^\Omega$, and $\Omega$ is the output domain of the competitors.
The competing players minimize their loss function $L_R, L_F$ respectively such that each player optimizes for itself but not for the group. To resolve this problem, our training cycle consists of two phases. In the first phase, we train the competitors by fixing the moderator network $M$ and minimizing
\begin{equation}
    E_{1} = \sum_i m \cdot L_R(R(\mathcal{D}_i)) + (1-m) \cdot L_F(F(\mathcal{D}_i)),
\end{equation}
where $\cdot$ is used to represent elementwise product throughout the paper.
However, the moderator $M$ also needs to be trained. This happens in the second phase of the training cycle. The competitors $\{R,F\}$ form a consensus and train the moderator $M$ such that it correctly distributes the data in the next phase of the training cycle. In the collaboration phase, we fix the competitors and train the moderator by minimizing,
\begin{equation}
    E_{2} = E_{1} + \sum_i L_M(\mathcal{D}_i, R, F)
\end{equation}
where $L_M$ is a loss that denotes a consensus between the competitors $\{R,F\}$. Competitive Collaboration can be applied to more general problems of training multiple task specific networks.
In the Appendix \ref{appendix:generalization}, we show the generalization of our method using an example of mixed domain learning on MNIST and SVHN digits, and analyze its convergence properties.

In the context of jointly learning depth, camera motion, optical flow and motion segmentation, the first player $R=(D,C)$ consists of the depth and camera motion networks that reason about the static regions in the scene. The second player $F$ is the optical flow network that reasons about the moving regions. For training the competitors, the motion segmentation
network $M$ selects networks $(D,C)$ on pixels that are static and selects $F$ on pixels that belong to moving regions. The competition ensures that $(D,C)$ reasons only about the static parts and prevents moving pixels from corrupting its training. Similarly, it prevents any static pixels from appearing in the training loss of $F$, thereby improving its performance in the moving regions. In the second phase of the training cycle, the competitors $(D,C)$ and $F$ now collaborate to reason about static scene and moving regions by forming a consensus that is used as a loss for training the moderator, $M$. In the rest of this section, we formulate the joint unsupervised estimation of depth, camera motion, optical flow and motion segmentation within this framework.
\paragraph{Notation.}
We use $\{D_\theta, C_\phi, F_\psi , M_\chi \}$, to denote the networks that estimate depth, camera motion, optical flow and motion segmentation respectively. The subscripts $\{\theta, \phi, \psi, \chi\}$ are the network parameters. We will omit the subscripts in several places for brevity. Consider an image sequence $I_-, I, I_+$ with target frame $I$ and temporally neighboring reference frames $I_-, I_+$. In general, we can have many neighboring frames. In our implementation, we use 5-frame sequences for $C_\phi$ and $M_\chi$ but for simplicity use 3 frames to describe our approach.
We estimate the depth of the target frame as
\begin{equation}
d = D_\theta(I).
\end{equation}
We estimate the camera motion, $e$, of each of the reference frames $I_-, I_+$ w.r.t.~the target frame $I$ as
\begin{equation}
e_{-}, e_{+} = C_\phi(I_-, I, I_+). \\
\end{equation}
Similarly, we estimate the segmentation of the target image into the static scene and moving regions. The optical flow of the static scene is defined only by the camera motion and depth. This generally refers to the structure of the scene. The moving regions have independent motion w.r.t. the scene. The segmentation masks corresponding to each pair of target and reference image are given by
\begin{equation}
m_{-}, m_{+} = M_\chi(I_-, I, I_+), \\
\end{equation}
where $m_{-}, m_{+} \in [0,1]^\Omega$
are the probabilities of regions being static in spatial pixel domain, $\Omega$.
Finally, the network $F_{\psi}$ estimates the optical flow. $F_{\psi}$ works with $2$ images at a time, and its weights are shared while estimating $u_-, u_+$, the backward and forward optical flow\footnote{Note that this is different from the forward and backward optical flow in the context of two-frame estimation.} respectively.
\begin{equation}
u_{-} = F_\psi(I, I_-), \qquad u_{+} = F_\psi(I, I_+).
\end{equation}
\paragraph{Loss.}
We learn the parameters of the networks $\{D_\theta, C_\phi, F_\psi , M_\chi \}$ by jointly minimizing the energy
\begin{equation}
\label{eq:loss}
E = \lambda_R E_R + \lambda_F E_F + \lambda_M E_M + \lambda_C E_C + \lambda_S E_S ,
\end{equation}
where $\{\lambda_R, \lambda_F, \lambda_M, \lambda_C, \lambda_S \}$ are the weights on the respective energy terms. The terms $E_R$ and $E_F$ are the objectives that are minimized by the two competitors reconstructing static and moving regions respectively. The competition for data is driven by $E_M$. A larger weight $\lambda_M$ will drive more pixels towards static scene reconstructor. The term $E_C$ drives the collaboration and $E_S$ is a smoothness regularizer.
The static scene term, $E_R$ minimizes the photometric loss on the static scene pixels given by
\begin{equation}
\label{eq:ER}
E_R = \sum_{s\in\{+,-\}} \sum_\Omega \rho \Big ( I, w_c(I_s, e_s, d) \Big )\cdot m_s
\end{equation}
where $\Omega$ is the spatial pixel domain, $\rho$ is a robust error function, and $w_c$ warps the reference frames towards the target frame according to depth $d$ and camera motion $e$.
Similarly, $E_F$ minimizes photometric loss on moving regions
\begin{equation}
\label{eq:EF}
E_F =  \sum_{s\in\{+,-\}} \sum_\Omega \rho  \Big( I , w_f(I_s, u_s) \Big) \cdot (1-m_s)
\end{equation}
where $w_f$ warps the reference image using flow $u$. We show the formulations for $w_c, w_f$ in the Appendix \ref{appendix:camera} and \ref{appendix:flow} respectively. We compute the robust error $\rho(x,y)$ as
\begin{equation}
{\scriptstyle
\rho (x,y) = \lambda_\rho \sqrt{(x\!-\!y)^2 \!+\! \epsilon^2} + (1 \!-\! \lambda_\rho) \Bigg[1 \!-\! \frac{(2 \mu_x \mu_y \!+\!c_1)(2\mu_{xy}\!+\!c_2)}{(\mu_x^2 \!+\! \mu_y^2 \!+\! c_1)(\sigma_x \!+\! \sigma_y \!+\! c_2)} \Bigg]
}
\end{equation}
\IGNORE{
\begin{equation}
\rho (x,y) = \lambda_\rho \sqrt{(x-y)^2 + \epsilon^2} + (1 - \lambda_\rho) \Bigg[1 - \frac{(2 \mu_x \mu_y +c_1)(2\mu_{xy}+c_2)}{(\mu_x^2 + \mu_y^2 + c_1)(\sigma_x + \sigma_y + c_2)} \Bigg]
\end{equation}
}
where $\lambda_\rho$ is a fixed constant and $\epsilon=0.01$. The second term is also known as the structure similarity loss~\cite{wang2004image} that has been used in previous work \cite{mahjourian2018googleicp,yin2018geonet}, and $\mu_x, \sigma_x$ are the local mean and variance over the pixel neighborhood with $c_1=0.01^2$ and $c_2=0.03^2$.

The loss $E_M$ minimizes the cross entropy, $H$, between the masks and a unit tensor regulated by $\lambda_M$
\begin{equation}
\label{eq:EM}
E_M =   \sum_{s\in\{+,-\}} \sum_\Omega H(\mathbf{1}, m_{s}) . \\
\end{equation}
A larger $\lambda_M$ gives preference to the static scene reconstructor $R$, biasing the scene towards being static.

Let $\nu(e,d)$  represent the optical flow induced by camera motion $e$ and depth $d$, as described in the Appendix \ref{appendix:camera}.
The consensus loss $E_C$ drives the collaboration and constrains the masks to segment moving objects by taking a consensus between flow of the static scene given by $\nu(e,d)$ and optical flow estimates from $F_\psi$. It is given by 
\begin{equation}
\label{eq:census}
E_C \!=\! \sum_{s\in\{\!+\!,\!-\!\}} \sum_{\Omega} H \big(\mathbb{I}_{\rho_R < \rho_F} \vee \mathbb{I}_{||\nu(e_s,d) \!- \! u_s|| \!<\! \lambda_c} , m_s\big)
\end{equation}
where $\mathbb{I} \in \{0,1\}$ is an indicator function  and equals $1$ if the condition in the subscript is true. The first indicator function favors mask assignments to the competitor that achieves lower photometric error on a pixel by comparing $\rho_R = \rho ( I, w_c(I_s, e_s, d) )$ and $\rho_F = \rho ( I, w_f(I_s, u_s) ) $. In the second indicator function, the threshold $\lambda_c$ forces $\mathbb{I} = 1$,
if the static scene flow $\nu(e,d)$ is close to the optical flow $u$, indicating a static scene. The symbol $\vee$ denotes logical OR between indicator functions. The consensus loss $E_C$ encourages a pixel to be labeled as static if $R$ has a lower photometric error than $F$ or if the induced flow of $R$ is similar to that of $F$.
Finally, the smoothness term $E_S$ acts as a regularizer on depth, segmentations and flow,
\begin{align}
E_S = \sum_\Omega ||\lambda_e \nabla d ||^2 + ||\lambda_e \nabla  u_- ||^2 + ||\lambda_e \nabla u_+ ||^2 \nonumber \\
+ ||\lambda_e \nabla m_- ||^2 + ||\lambda_e \nabla m_+ ||^2 ,
\end{align}
where $\lambda_e = e^{-\nabla I}$ (elementwise) and $\nabla$ is the first derivative along spatial directions \cite{rother2004grabcut}. The term $\lambda_e$ ensures that smoothness is guided by edges of the images.

\paragraph{Inference.}
The depth $d$ and camera motion $e$ are directly inferred from network outputs. The motion segmentation $m^*$ is obtained by the output of mask network $M_\chi$ and the consensus between the static flow and optical flow estimates from $F_\chi$. It is given by
\begin{equation}
m^* = \mathbb{I}_{m_+ \cdot m_- > 0.5} \vee \mathbb{I}_{||\nu(e_+,d) - u_+|| < \lambda_c}.
\label{eqn:jointmask}
\end{equation}
The first term takes the intersection of mask probabilities inferred by $M_\chi$ using forward and backward reference frames. The second term takes a consensus between flow estimated from $R=(D_\theta,C_\phi)$ and $F_\psi$ to reason about the masks. The final masks are obtained by taking the union of both terms.
Finally, the full optical flow, $u^*$, between $(I,I_+)$ is a composite of optical flows from the static scene and the independently moving regions given by
\begin{equation}
\label{eq:of}
u^* = \mathbb{I}_{m^* > 0.5} \cdot  \nu(e_+,d) + \mathbb{I}_{m^* \leq 0.5} \cdot u_+.
\end{equation}
The loss in  Eq. (\ref{eq:loss}) is formulated to minimize the reconstruction error of the neighboring frames. Two competitors, the static scene reconstructor $R=(D_\theta, C_\phi)$ and moving region reconstructor $F_\psi$ minimize this loss. The reconstructor $R$ reasons about the static scene using Eq. (\ref{eq:ER}) and the reconstructor $F_\psi$ reasons about the moving regions using Eq.~(\ref{eq:EF}).  The moderation is achieved by the mask network, $M_\chi$ using Eq. (\ref{eq:EM}). Furthermore, the collaboration between $R,F$ is driven using Eq.~(\ref{eq:census}) to train the network $M_\chi$.

If the scenes are completely static, and only the camera moves, the mask forces $(D_\theta, C_\phi)$ to reconstruct the  whole scene. However, $(D_\theta, C_\phi)$ are  wrong in the independently moving regions of the scene, and these regions are reconstructed using $F_\psi$.
The moderator $M_\chi$ is trained to segment static and moving regions correctly  by taking a consensus from $(D_\theta, C_\phi)$ and $F_\psi$ to reason about static and moving parts on the scene, as seen in Eq. (\ref{eq:census}).
Therefore, our training cycle has two phases. In the first phase, the moderator $M_\chi$ drives competition between two models $(D_\theta, C_\phi)$ and $F_\psi$ using Eqs.~(\ref{eq:ER}, \ref{eq:EF}). In the second phase, the competitors $(D_\theta, C_\phi)$ and $F_\psi$ collaborate together to train the moderator $M_\chi$ using Eqs.~(\ref{eq:EM},\ref{eq:census}).

\newcommand{\visualresultwidth}{0.2\textwidth}
\begin{figure*}[t]
  \centering
  \centerline{
    \includegraphics[width=\visualresultwidth]{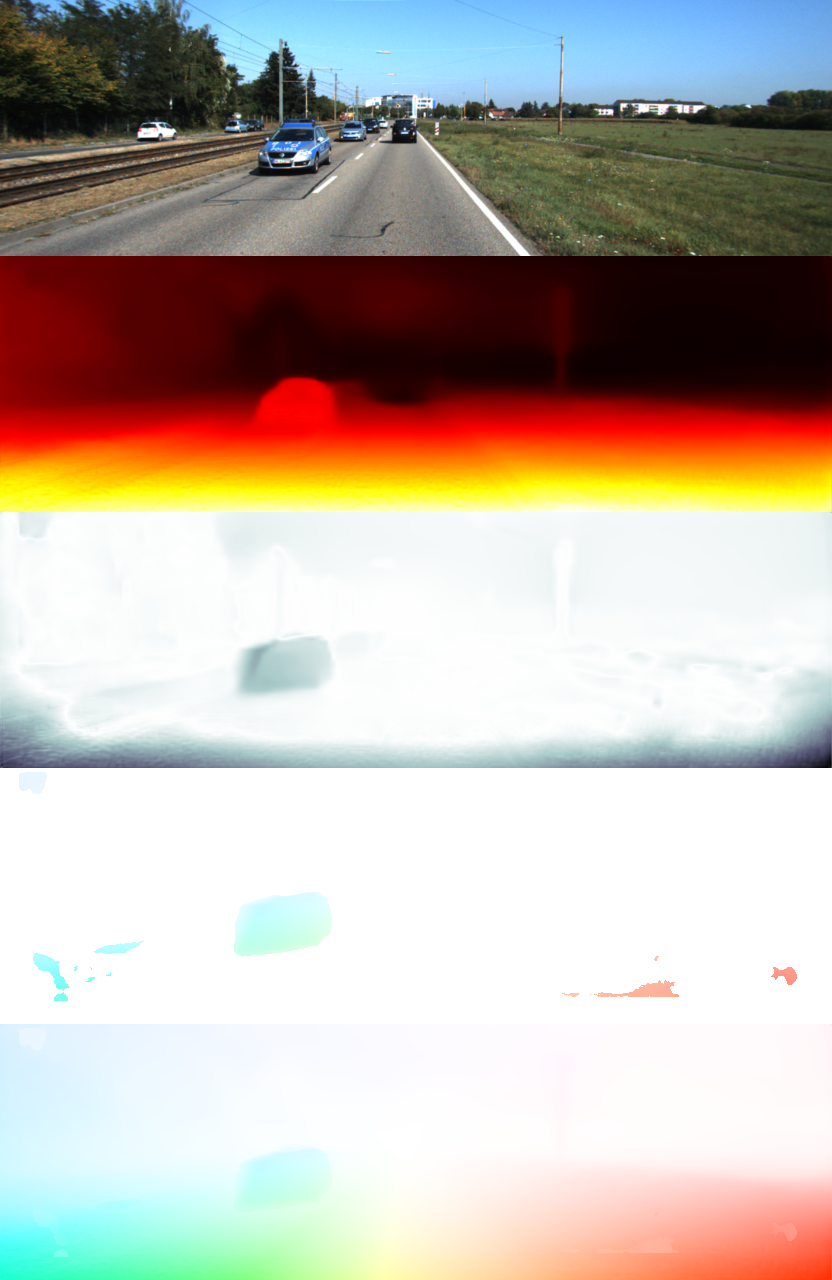}%
    \includegraphics[width=\visualresultwidth]{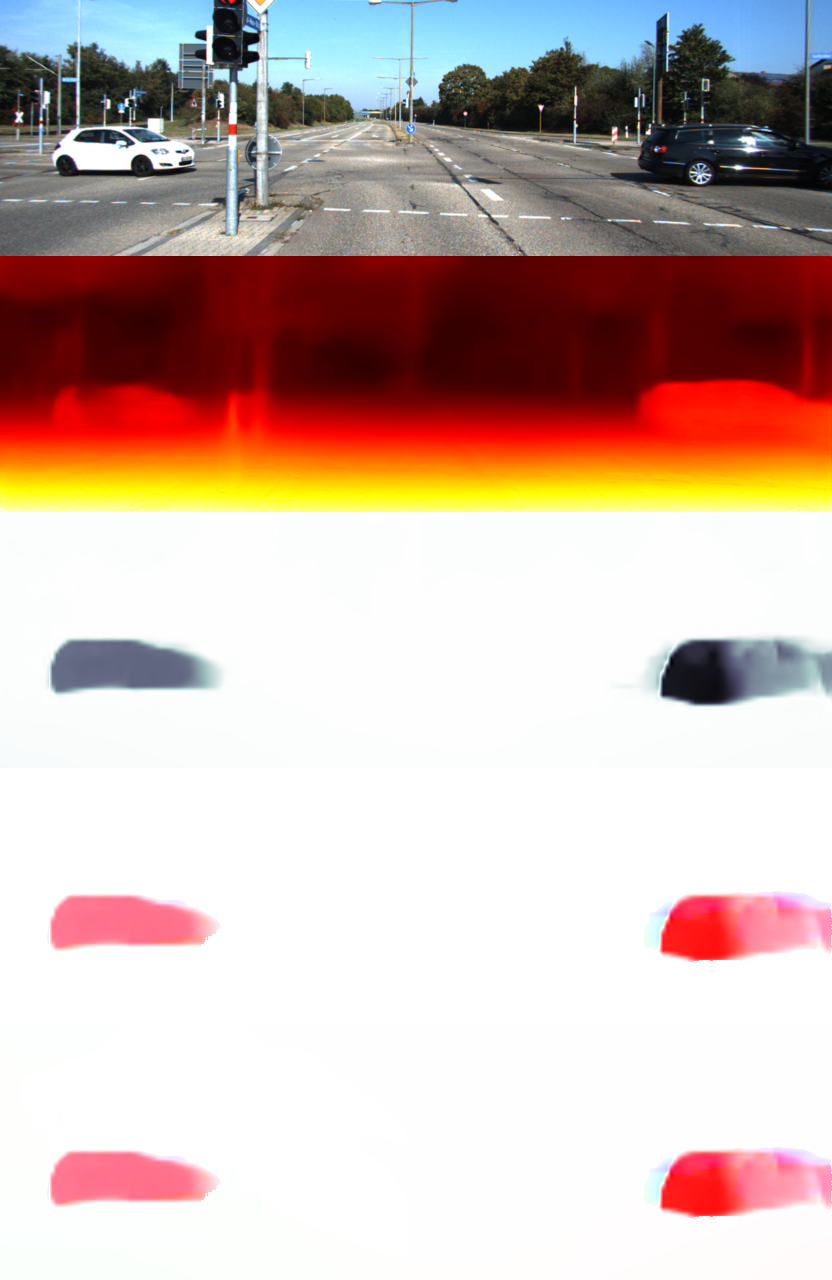}%
    \includegraphics[width=\visualresultwidth]{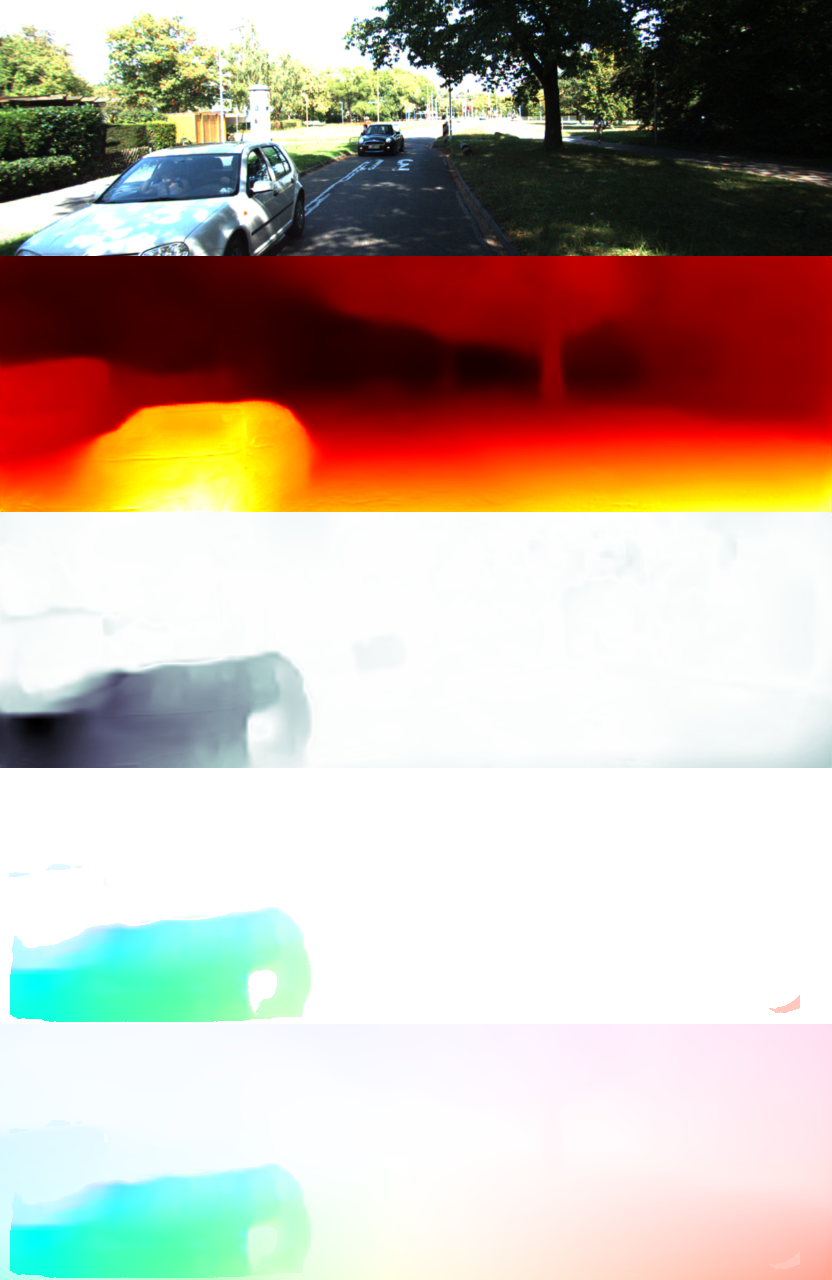}%
    \includegraphics[width=\visualresultwidth]{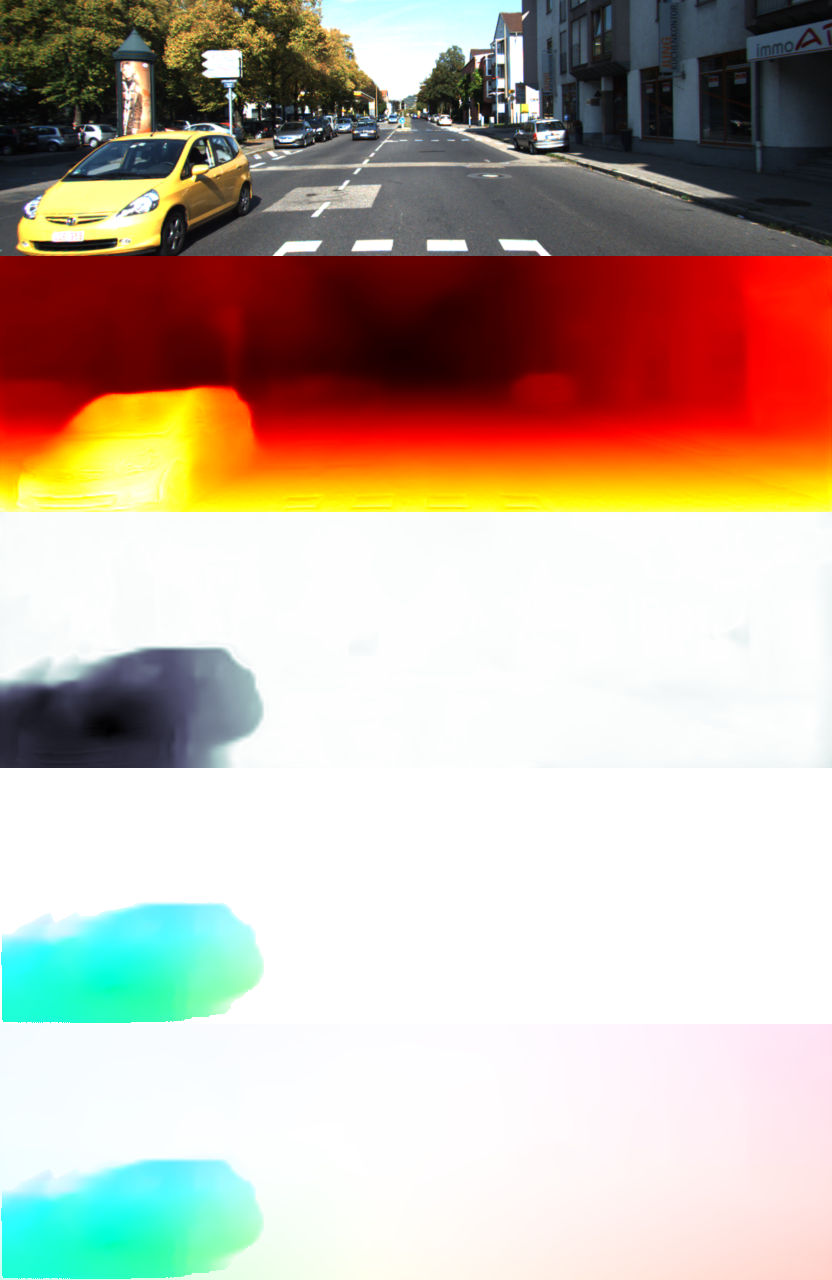}%
    \includegraphics[width=\visualresultwidth]{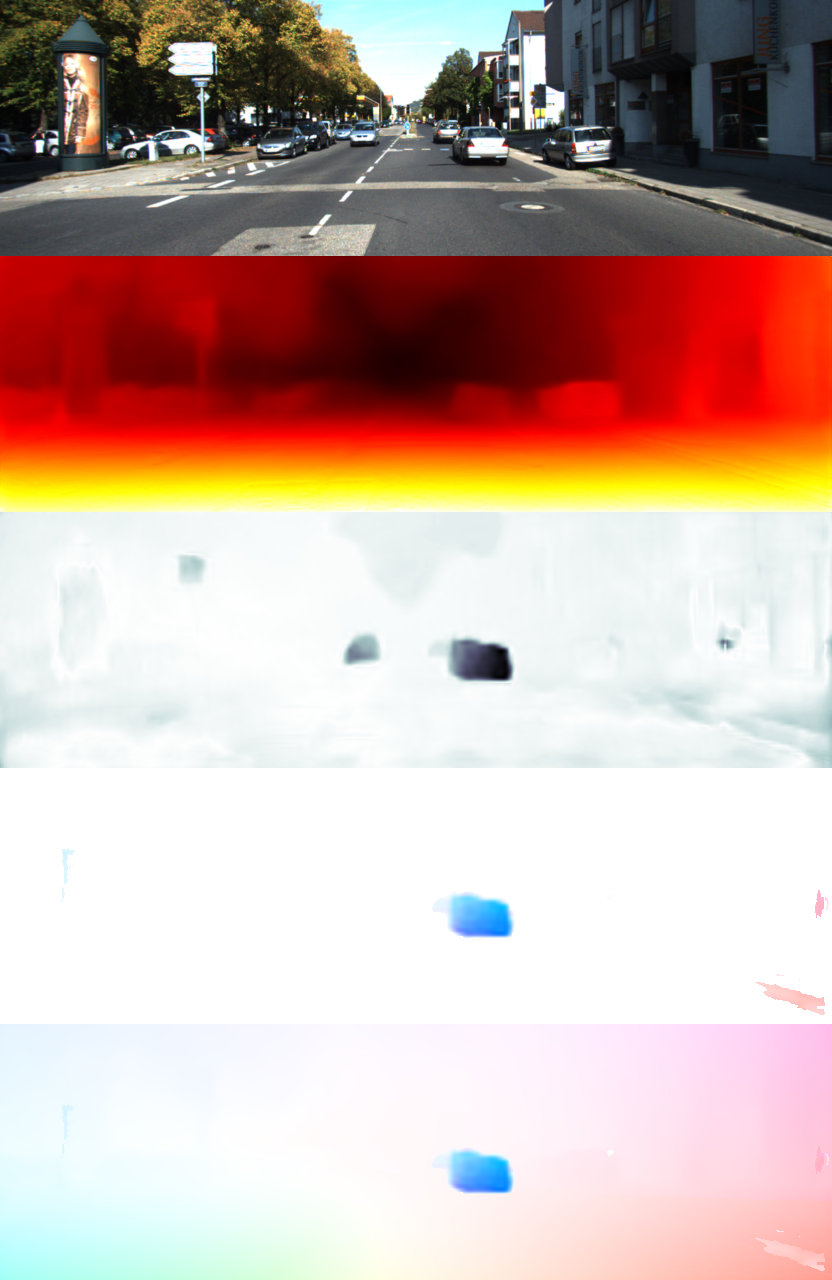}
    }
    \vspace{-0.1cm}
	\caption{ {\bf Visual results.} Top to bottom: Sample image, estimated depth, soft consensus masks, motion segmented optical flow and combined optical flow. }
    \vspace{-0.5cm}
    \label{fig:viz}
\end{figure*}

\SetKwFor{Competition}{Competition Step}{}{}
\SetKwFor{Collaboration}{Collaboration Step}{}{}
\SetKwFor{Loop}{Loop}{}{EndLoop}
\begin{algorithm}[b]
\vspace{-0.2cm}
\SetAlgoLined
\KwResult{Trained Network Parameters, $(\theta, \phi, \psi, \chi)$}
Define $\lambda=(\lambda_R, \lambda_F, \lambda_M, \lambda_C)$\;
 Randomly initialize $(\theta, \phi, \psi, \chi)$\;
 Update $(\theta, \phi)$ by jointly training $(D_\theta, C_\phi)$ with $\lambda=(1.0,0.0,0.0,0.0)$\;
 Update $\psi$ by training $F_\psi$ with  $\lambda=(0.0,1.0,0.0,0.0)$\;
 Update $\chi$ by jointly training $(D_\theta, C_\phi, F_\psi, M_\chi)$ with $\lambda=(1.0,0.5,0.0,0.0)$\;
  \Loop{}{ \Competition{}{
  Update $\theta, \phi$ by jointly training $(D_\theta, C_\phi,$  $F_\psi, M_\chi)$ with $\lambda=(1.0,0.5,0.05,0)$ \;
  Update $\psi$ by jointly training $(D_\theta, C_\phi, F_\psi, M_\chi)$ with $\lambda=(0.0,1.0,0.005,0)$  \;}
  \Collaboration{}{
  Update $\chi$ by jointly training $(D_\theta, C_\phi, F_\psi, M_\chi)$ with $\lambda=(1.0,0.5,0.005,0.3)$ \;}}
 \caption{Network Training Algorithm}
 \label{algo:training}
\end{algorithm}

\section{Experiments}
\paragraph{Network Architecture.}

For the depth network, we experiment with DispNetS~\cite{zhou2017unsupervised} and DispResNet where we replace convolutional blocks with residual blocks~\cite{he2016resnet}. The network $D_\theta$ takes a single RGB image as input and outputs depth. For the flow network, $F_\psi$, we experiment with both FlowNetC \cite{dosovitskiy2015flownet} and PWC-Net~\cite{sun2017pwc}. The PWC-Net uses the multi-frame unsupervised learning framework from Janai et al.~\cite{janai2018unsupervised}. The network $F_\psi$ computes optical flow between a pair of frames. The networks $C_\phi, M_\chi$ take a 5 frame sequence $(I_{--},I_-,I,I_+, I_{++})$ as input.
The mask network $M_\chi$ has an encoder-decoder architecture. The encoder consists of stacked residual convolutional layers. The decoder has stacked upconvolutional layers to produce masks $(m_{--},m_-, m_+, m_{++})$ of the reference frames. The camera motion network $C_\phi$ consists of stacked convolutions followed by adaptive average pooling of feature maps to get the camera motions $(e_{--}, e_-, e_+, e_{++})$. The networks $D_\theta,F_\psi, M_\chi$ output their results at 6 different spatial scales. The predictions at the finest scale are used.
The highest scale is of the same resolution as the image, and each lower scale reduces the resolution by a factor of 2. We show the network architecture details in the Appendix \ref{appendix:nets}.
\vspace{-0.3cm}
\paragraph{Network Training.}
We use raw KITTI sequences \cite{Geiger2012CVPR} for training using Eigen et al.'s split \cite{eigen2014depth} that is consistent across related works \cite{eigen2014depth,liu2016learning,mahjourian2018googleicp,yin2018geonet,zhou2017unsupervised,zou2018df}. We train the networks with a batch size of 4 and learning rate of $10^{-4}$ using ADAM~\cite{kingma2014adam} optimization. The images are scaled to $256 \times 832$ for training. The data is augmented with random scaling, cropping and horizontal flips. We use Algorithm \ref{algo:training} for training. Initially, we train $(D_\theta, C_\phi)$ with only photometric loss over static pixels $E_R$ and smoothness loss $E_S$ while other loss terms are set to zero. Similarly, we train $F_\psi$ independently with photometric loss over all pixels and smoothness losses. The models $(D_\theta, C_\phi),F_\psi$ at this stage are referred to as `basic' models in our experiments. We then learn $M_\chi$ using the joint loss. We use $\lambda_R=1.0, \lambda_F=0.5$ for joint training because the static scene reconstructor $R$ uses 4 reference frames in its loss, whereas the optical flow network $F$ uses 2 frames. Hence, these weights normalize the loss per neighboring frame. We iteratively train $(D_\theta,C_\phi), F_\psi,M_\chi$ using the joint loss while keeping the other network weights fixed. The consensus weight $\lambda_C=0.3$ is used only while training the mask network. Other constants are fixed with $\lambda_S=0.005$, and threshold in Eq.(\ref{eqn:jointmask}), $\lambda_c=0.001$. The constant $\lambda_\rho=0.003$ regulates the SSIM loss and is chosen empirically. We iteratively train the competitors $(D_\theta,C_\phi), F_\psi$ and moderator $M_\chi$ for about 100,000 iterations at each step until validation error saturates.

\begin{table*}[t]
\begin{center}
\begin{tabular}{lccccc|ccc}
& & \multicolumn{4}{c|}{Error} & \multicolumn{3}{c}{Accuracy, $\delta$ } \\ \cline{3-6}  \cline{7-9}
Method& Data &AbsRel&SqRel&RMS&RMSlog &$<$$1.25$&$<$$1.25^2$& $<$$1.25^3$\\ \Xhline{3\arrayrulewidth}
{ Eigen et al.~\cite{eigen2014depth} coarse}& k & 0.214& 1.605& 6.563& 0.292& 0.673& 0.884 &0.957\\
{ Eigen et al.~\cite{eigen2014depth} fine}& k & 0.203 &1.548& 6.307& 0.282& 0.702& 0.890& 0.958 \\
{ Liu et al. \cite{liu2016learning}} & k & 0.202& 1.614& 6.523& 0.275& 0.678& 0.895& 0.965\\ \hline
{ Zhou et al.~\cite{zhou2017unsupervised}} &cs+k &0.198 & 1.836 & 6.565 & 0.275 & 0.718 & 0.901 & 0.960 \\
{ Mahjourian et al.~\cite{mahjourian2018googleicp} }&cs+k & 0.159 & 1.231 & 5.912 & 0.243 & 0.784 & 0.923 & 0.970 \\
{ Geonet-Resnet \cite{yin2018geonet} }&cs+k &{0.153} & 1.328 & 5.737 & 0.232 & 0.802 & {0.934} & 0.972 \\
{ DF-Net \cite{zou2018df} }&cs+k & 0.146 & 1.182 & 5.215 & \textbf{0.213} & 0.818 & \textbf{0.943} & \textbf{0.978}\\
{ CC (ours)} & cs+k & \textbf{0.139} & \textbf{1.032} & \textbf{5.199} & \textbf{0.213} & \textbf{0.827} & \textbf{0.943}&  { 0.977} \\ 
\hline
{ Zhou et al.*~\cite{zhou2017unsupervised}} &k &0.183 & 1.595 & 6.709 & 0.270 & 0.734 & 0.902 & 0.959 \\
{ Mahjourian et al.~\cite{mahjourian2018googleicp}}&k & 0.163 & 1.240 & 6.220 & 0.250 & 0.762 & 0.916 & 0.968 \\
{ Geonet-VGG \cite{yin2018geonet} }&k &{0.164} & 1.303 & 6.090 & 0.247 & {0.765} & {0.919} & {0.968} \\
{ Geonet-Resnet \cite{yin2018geonet} }&k &{0.155} & 1.296 & 5.857 & 0.233 & {0.793} & {0.931} & {0.973} \\
{ Godard et el.~\cite{godard2018digging} } & k & 0.154 & 1.218 & 5.699 & 0.231 & 0.798 & 0.932 & 0.973 \\
{ DF-Net \cite{zou2018df} }&k & 0.150  &1.124&  5.507& 0.223& 0.806& 0.933& 0.973 \\

{ CC (ours)} & k &  \textbf{0.140}&     \textbf{1.070}&     \textbf{5.326}&     \textbf{0.217}&     \textbf{0.826}&     \textbf{0.941}&    \textbf{0.975} %
\end{tabular}
\caption{Results on Depth Estimation. Supervised methods are shown in the first rows. Data refers to the training set: Cityscapes (cs) and KITTI (k). Zhou el al.*~shows improved results from their github page. }
\label{tab:depth}
\end{center}
\vspace{-0.4cm}
\end{table*}

\begin{table*}
\begin{center}
\begin{tabular}{lccccccc|ccc}
& & & &\multicolumn{4}{c|}{Error} & \multicolumn{3}{c}{Accuracy, $\delta$ } \\ \cline{5-8}  \cline{9-11}
Method& Data &Net $D$&Net $F$ &AbsRel&SqRel&RMS&RMSlog &$<$$1.25$&$<$$1.25^2$& $<$$1.25^3$\\ \Xhline{3\arrayrulewidth}

Basic &k& DispNet & - & 0.184 & 1.476 & 6.325 & 0.259 & 0.732 & 0.910 & 0.967 \\
Basic + ssim &k& DispNet & - & 0.168 & 1.396 & 6.176 & 0.244 & 0.767 & 0.922 & 0.971 \\
CC + ssim & k & DispNet & FlowNetC &
{0.148}  &  {1.149} & {5.464} & {0.226} & {0.815} &{0.935} & {0.973} \\
CC + ssim & k & DispResNet & FlowNetC &  0.144 &1.284 & 5.716 & 0.226 & 0.822 & 0.938 & 0.973 \\
CC + ssim & k & DispResNet & PWC Net &  {0.140}&    {1.070}&    {5.326}&     {0.217}&  {0.826}& {0.941}& {0.975} \\
CC + ssim & cs+k & DispResNet & PWC Net & \textbf{0.139} & \textbf{1.032} & \textbf{5.199} & \textbf{0.213} & \textbf{0.827} & \textbf{0.943}&  \textbf{0.977}

\end{tabular}
\caption{Ablation studies on Depth Estimation. Joint training using Competitive Collaboration and better architectures improve the results. The benefits of CC can be seen when depth improves by using a better network for flow (row 4 vs 5).}
\label{tab:depth_ablation}
\end{center}
\vspace{-0.4cm}
\end{table*}

\begin{table}
\begin{center}
\begin{tabular}{lcc}
Method   &  Sequence 09 & Sequence 10\\ \Xhline{3\arrayrulewidth}
{ORB-SLAM (full)} & {\small 0.014 $\pm$ 0.008}  & {\small 0.012 $\pm$ 0.011  }\\
{ORB-SLAM (short)} & {\small 0.064 $\pm$ 0.141} &{\small 0.064 $\pm$ 0.130 }\\
{Mean Odometry} & {\small 0.032 $\pm$ 0.026} & {\small 0.028 $\pm$ 0.023 } \\
{Zhou et al.~\cite{zhou2017unsupervised}} & {\small 0.016 $\pm$ 0.009} &{\small 0.013 $\pm$ 0.009}  \\
{Mahjourian et al.~\cite{mahjourian2018googleicp}}& {\small 0.013 $\pm$ 0.010} & {\small 0.012 $\pm$ 0.011 } \\
{Geonet \cite{yin2018geonet}} &{\textbf{\small 0.012 $\pm$ 0.007}}&{\small 0.012 $\pm$ 0.009}  \\
{DF-Net \cite{zou2018df} } & {\small 0.017 $\pm$ 0.007} & {\small 0.015 $\pm$ 0.009} \\
{Basic (ours)} & {\small 0.022 $\pm$ 0.010} & {\small 0.018 $\pm$ 0.011}  \\
{Basic + ssim (ours)} & {\small 0.017 $\pm$ 0.009} & {\small 0.015 $\pm$ 0.009}  \\
{CC + ssim (ours)} & {\textbf{\small 0.012 $\pm$ 0.007}} & {\textbf{\small 0.012 $\pm$ 0.008}}  \\
\end{tabular}
\caption{Results on Pose Estimation.}
\label{tab:pose}
\vspace{-0.4cm}
\end{center}
\end{table}

\paragraph{Monocular Depth and Camera Motion Estimation.} We obtain state of the art results on single view depth prediction and camera motion estimation as shown in Tables \ref{tab:depth} and \ref{tab:pose}. The depth is evaluated on the Eigen et al.~\cite{eigen2014depth} split of the raw KITTI dataset \cite{Geiger2012CVPR} and camera motion is evaluated on the KITTI Odometry dataset \cite{Geiger2012CVPR}. These evaluation frameworks are consistent with previous work \cite{eigen2014depth,liu2016learning,mahjourian2018googleicp,yin2018geonet}. All depth maps are capped at 80 meters. As shown in Table \ref{tab:depth}, by training our method only on KITTI \cite{Geiger2012CVPR}, we get similar or better performance than competing methods like \cite{yin2018geonet, zou2018df} that use a much bigger Resnet-50 architecture \cite{he2016resnet} and are trained on the larger Cityscapes dataset \cite{Cordts2016Cityscapes}. Using Cityscapes in our training further improves our performance on depth estimation benchmarks (cs+k in Table \ref{tab:depth}).

Ablation studies on depth estimation are shown in Table \ref{tab:depth_ablation}. In the basic mode, our network architecture, DispNet for depth and camera motion estimation is most similar to \cite{zhou2017unsupervised} and this is reflected in the performance of our basic model. We get some performance improvements by adding the SSIM loss \cite{wang2004image}. However, we observe that using the Competitive Collaboration (CC) framework with a joint loss results in larger performance gains in both tasks. Further improvements are obtained by using a better network architecture, DispResNet. Greater improvements in depth estimation are obtained when we use a better network for flow, which shows that improving on one task improves the performance of the other in the CC framework (row 4 vs 5 in Table \ref{tab:depth_ablation}).

The camera motion estimation also shows similar performance trends as shown in Table \ref{tab:pose}. Using a basic model, we achieve similar performance as the baseline \cite{zhou2017unsupervised}, which improves with the addition of the SSIM loss. Using the CC framework leads to further improvements in performance.

In summary, we show that joint training using CC boosts performance of single view depth prediction and camera motion estimation. We show qualitative results in Figure~\ref{fig:viz}. In the Appendix, we show additional evaluations using Make3D dataset \cite{saxena2006learning} (\ref{appendix:extra}) and more qualitative results (\ref{appendix:results}).
\vspace{-0.2cm}

\begin{table}[t]
\begin{center}
\begin{tabular}{lccc}
 &  \multicolumn{2}{c}{ Train } & Test  \\ \cline{2-4}
{Method} & {EPE} & {Fl} & {Fl} \\ \Xhline{3\arrayrulewidth}
{FlowNet2 \cite{ilg2017flownet} } & 10.06 & 30.37 \% & -\\
{SPyNet \cite{spynet} } & 20.56 & 44.78\% & -\\
\hline
{UnFlow-C \cite{meister2017unflow}} & 8.80 & 28.94\% &   29.46\% \\
{UnFlow-CSS \cite{meister2017unflow}} & 8.10 & \textbf{23.27\% }& {- } \\
{Back2Future \cite{janai2018unsupervised} } & \textbf{6.59} & - & \textbf{22.94\%} \\
{Back2Future* \cite{janai2018unsupervised} } & {7.04} &  {24.21\%} & - \\\hline
{Geonet \cite{yin2018geonet}} &  10.81 & - & {-} \\
{DF-Net \cite{zou2018df} } &8.98 &26.01\%  &25.70\% \\
{CC (ours)} & 6.21 & 26.41\% & -\\
{CC-uft (ours)} & \textbf{5.66} & \textbf{20.93\%} & \textbf{ 25.27\%} \\
\end{tabular}
\caption{Results on Optical Flow.  We also compare with supervised methods (top 2 rows) that are trained on synthetic data only; unsupervised methods specialized for optical flow (middle 3 rows) and joint methods that solve more than one task (bottom 4 rows). * refers to our Pytorch implementation used in our framework which gives slightly lower accuracy.}
\label{tab:flow}
\end{center}
\vspace{-0.4cm}
\end{table}

\begin{table}[t]
\begin{center}
\begin{tabular}{lccccc}
 & & & \multicolumn{3}{c}{ Average EPE }  \\ \cline{4-6}
{\small Method} & Net D & Net F & {SP} & {MP} & {Total} \\ \Xhline{3\arrayrulewidth}
{\small $R$} & {\small DispNet} & -& {\small 7.51} & {\small 32.75} & {\small 13.54} \\
{\small $F$} & - & {\small FlowNetC} & {\small 15.32} & {\small 6.20} & {\small 14.68} \\
{\small CC} & {\small DispNet} & {\small FlowNetC} & {\small 6.35} & {\small 6.16} & {\small 7.76} \\
 {\small 11.86}  \\
{\small CC} & {\small DispResNet} & {\small PWC Net} & {\small 5.67} & {\small 5.04} & {\small 6.21} \\
\end{tabular}
\caption{Ablation studies on Flow estimation. SP, MP refer to static scene and moving region pixels. EPE is computed over KITTI 2015 training set. $R,F$ are trained independently without CC. }
\label{tab:flow_ablation}
\end{center}
\vspace{-0.4cm}
\end{table}
\vspace{-0.4cm}

\paragraph{Optical Flow Estimation.}
We compare the performance of our approach with competing methods using the KITTI 2015 training set \cite{Geiger2012CVPR} to be consistent with previous work \cite{meister2017unflow, yin2018geonet}. We obtain state of the art performance among joint methods as shown in Table \ref{tab:flow}. Unsupervised fine tuning (CC-uft) by setting $\lambda_M=0.02$ gives more improvements than CC as masks now choose best flow between $R$ and $F$ without being overconstrained to choose $R$. In contrast, UnFlow-CSS \cite{meister2017unflow} uses 3 cascaded networks to refine optical flow at each stage. Geonet \cite{yin2018geonet} and DF-Net \cite{zou2018df} are more similar to our architecture but use a larger ResNet-50 architecture. Back2Future~\cite{janai2018unsupervised} performs better than our method in terms of outlier error, but not in terms of average end point error due to better occlusion reasoning.

In Table \ref{tab:flow_ablation}, we observe that by training the static scene reconstructor $R$ or moving region reconstructor $F$ independently leads to worse performance. This happens because $R$ can not reason about dynamic moving objects in the scene. Similarly $F$ is not as good as $R$ for reasoning about static parts of the scene, especially in occluded regions. Using them together, and compositing the optical flow from both as shown in Eq.~(\ref{eq:of}) leads to a large improvement in performance. Moreover, using better network architectures further improves the performance under CC framework. We show qualitative results in Figure~\ref{fig:viz} and in the Appendix \ref{appendix:results}.

\vspace{-0.4cm}
\paragraph{Motion Segmentation.}
We evaluate the estimated motion segmentations using the KITTI 2015 training set \cite{Geiger2012CVPR} that provides ground truth segmentation for moving cars.
Since our approach does not distinguish between different semantic classes while estimating segmentation, we evaluate segmentations only on car pixels. Specifically, we only consider car pixels and compute Intersection over Union (IoU) scores for moving and static car pixels. In Table~\ref{tab:motionseg}, we show the IoU scores of the segmentation masks obtained using our technique under different conditions.
We refer to the masks obtained with the motion segmentation network $\big ( \mathbf{I}_{m_-m_+ > 0.5} \big ) $ as `MaskNet' and refer to the
masks obtained with flow consensus $\big ( \mathbf{I}_{||\nu(e_+,d) - u_+|| < \lambda_c} \big )$ as `Consensus'. The final motion segmentation
masks $m^*$ obtained with the intersection of the above two estimates are referred to as `Joint'
(Eq.~\ref{eqn:jointmask}). IoU results indicate
substantial IoU improvements with `Joint' masks compared to both `MaskNet' and `Consensus' masks, illustrating the complementary nature
of different masks. Qualitative results are shown in Figure \ref{fig:viz} and in the Appendix \ref{appendix:results}.

\begin{table}
\begin{center}
  \begin{tabular}{lccc}
    \textbf{} & {Overall} & {Static Car} & {Moving Car}\\
    \midrule
  MaskNet & 41.64 & 30.56  & 52.71\\
  Consensus &  51.52 & 47.30 & 55.74\\
  Joint &  \textbf{56.94}   & \textbf{55.77}     & \textbf{58.11}\\
\end{tabular}
\caption{Motion Segmentation Results. Intersection Over Union (IoU) scores on
KITTI2015 training dataset images computed over car pixels.}
\label{tab:motionseg}
\end{center}
\end{table}

\section{Conclusions and Discussion}

Typically, learning to infer depth from a single image requires training images with ground truth depth scans, and
learning to compute optical flow relies on synthetic data, which may not generalize to real image sequences.
For static scenes, observed by a moving camera, these two problems are related by camera motion; depth and camera motion completely determine the 2D optical flow.
This holds true over several frames if the scene is static and only the camera moves.
Thus by combining depth, camera, and flow estimation, we can learn single-image depth by using information from several frames
during training.
This is particularly critical for unsupervised training since
both depth and optical flow are highly ill-posed.
Combining evidence from multiple tasks and multiple frames helps to synergistically constrain the problem.
This alone is not enough, however, as real scenes contain multiple moving objects that do not conform to static scene geometry.
Consequently, we also learn to segment the scene into static and moving regions without supervision.
In the independently moving regions, a generic flow network learns to estimate the optical flow.

To facilitate this process we introduce Competitive Collaboration in which networks both compete and cooperate.
We demonstrate that this results in top performance among unsupervised methods for all subproblems.
Additionally, the moderator learns to segment the scene into static and moving regions without any direct supervision.

\textbf{Future Work.}
We can add small amounts of supervised training, with which we expect to significantly boost performance on benchmarks, cf.~\cite{meister2017unflow}.
We could use, for example, sparse depth and flow from KITTI and segmentation from Cityscapes to selectively provide ground truth to different networks.
A richer segmentation network
together with semantic segmentation should improve non-rigid segmentation.
For automotive applications, the depth map formulation should be extended to a world coordinate system, which would support the integration of depth information over long image sequences.
Finally, as shown in \cite{Wulff17cvpr}, the key ideas of using layers and geometry apply to general scenes beyond the automotive case and we should be able to train this method to work with generic scenes and camera motions.

\subsection*{Acknowledgements}
We thank Frederik Kunstner for verifying the proofs. We are grateful to Cl\'ement Pinard for his github repository. We use it as our initial code base. We thank Georgios Pavlakos for helping us with several revisions of the paper. We thank Joel Janai for preparing optical flow visualizations, and Cl\'ement Gorard for his Make3d evaluation code.

{\small
\bibliographystyle{ieee}
\bibliography{egbib}
}
\begin{appendices}
\section{Appendix}

\subsection{Competitive Collaboration as a General Learning Framework}
\label{appendix:generalization}

Competitive collaboration (CC) can be seen as a general learning
framework for training multiple task-specific networks.
To showcase this generality, we demonstrate CC on a
mixed-domain
classification problem in Section~\ref{appendix:mnist} and
analyze CC convergence properties in Section~\ref{appendix:convergence}.

\subsubsection{Mixed Domain Classification}
\label{appendix:mnist}

Digit classification is the task of classifying a given image $I$ into one of the 10 digit classes $t \in \{0,1,2, ..,9\}$.
Two most widely used datasets for digit classification include images of the postal code digits, MNIST \cite{Lecun:1998:MNist} and street view house numbers, SVHN \cite{Netzer:2011:SVHN}. For our setup, we take the samples from both of the datasets, and shuffle them together. This means that, although an image and a target, $(I_i,t_i)$ form  a pair, there is no information if the digits came from MNIST or SVHN.

We now train our model under Competitive Collaboration framework given the mixed-domain dataset MNIST+SVHN, a mixture of MNIST and SVHN. The model consists of two networks $R_x$ and $F_x$ that compete with each other regulated by a moderator $M_y$ which assigns training data to each of the competitors. Here, $x$ denotes the combined weights of the two competitor networks $(R,F)$ and $y$ denotes the weight of the moderator network $M$. The networks are trained using an alternate optimization procedure consisting of two phases. In the competition phase, we train the competitors by fixing the moderator $M$ and minimizing,
\begin{equation}
    E_1 = \sum_i m_i \cdot H(R_x(I_i), t_i) + (1-m_i) \cdot H(F_x(I_i), t_i)
\end{equation}
where $m_i = M_y(I_i) \in [0,1]$ is the output of the moderator and is the probability of assigning a sample to $R_x$. $H(R_x(I_i), t_i)$ is the cross entropy classification loss on the network $R_x$ and a similar loss is applied on network $F_x$.

During the collaboration phase, we fix the competitors and train the moderator by minimizing,
\begin{equation}
\begin{split}
& E_2 = E_1 + \\
& \sum_i \lambda \cdot \begin{cases}
- \log( m_i + \varepsilon) & \text{if } L_{R_i} < L_{F_i},\\
- \log(1 - m_i + \varepsilon) & \text{if } L_{R_i} \geq L_{F_i}.
\end{cases}
\end{split}
\end{equation}

where $L_{R_i} = H(R_x(I_i), t_i) $ is the cross entropy loss from network $R_x$ and similarly $L_{F_i} = H(F_x(I_i), t_i) $.
In addition to the above loss function $E_1$, we use an additional
constraint on the moderator output that encourages the variance of $m$, $\sigma_m^2 = \Sigma_i (m_i - \bar{m})^2 $ to be high, where $\bar{m}$ is the mean of $m$ within a batch. This encourages the moderator to assign images to both the models, instead of always assigning them to a single model.

In an ideal case, we expect the moderator to correctly classify MNIST digits from SVHN digits. This would enable each of the competitors to specialize on either MNIST or SVHN, but not both. In such a case, the accuracy of the model under CC would be better than training a single network on the MNIST+SVHN mixture.

\paragraph{Experimental Results} For simplicity, we use a CNN with 2 convolutional layers followed by 2 fully-connected layers for both the digit classification networks $(R,F)$ as well as the moderator network $M$. Each of the convolutional layers use a kerel size of 5 and 40 feature maps. Each of the fully connected layers have 40 neurons.

We now compare the performance of the CC model on MNIST+SVHN mixture with training a single network on the same dataset. We see that our performance is better on the mixture dataset as well as individual datasets (see Table~\ref{tab:mnist_svhn}). As shown in Table~\ref{tab:mnist_svhn}, the network $R$ specializes on SVHN digits and network $F$ specializes on MNIST digits. By using the networks $(R,F,M)$, we get the best results as $M$ picks the specialized networks depending on the data sample.

We also examine the classification accuracy of the moderator on MNIST and SVHN digits. We observe that moderator can accurately classify the digits into either MNIST or SVHN without any labels
(see Table \ref{tab:mod_accuracy}).
The moderator learns to assign 100$\%$ of MNIST digits to $F$ and 100$\%$ of SVHN digits to $R$. This experiment provides further evidence to support the notion that CC can be generalized to other problems.

\begin{table}
\begin{center}
  \begin{tabular}{lcccc}
    \textbf{} & {Training} & {M} & {S} & {M+S}\\
    \midrule
$R$  &  Basic & 1.34 & 11.88 & 8.96\\ \hline
$R$ & CC   & 1.41 & \textbf{11.55}  & 8.74 \\
$F$ & CC  & \textbf{1.24} & 11.75  & 8.84  \\
$R,F,M$ & CC  &  \textbf{1.24} & \textbf{11.55} & \textbf{8.70}\\
  \bottomrule
\end{tabular}
\caption{Percentage classification errors. M and S refer to MNIST and SVHN respectively.}
\label{tab:mnist_svhn}
\end{center}
\end{table}

\begin{table}
\begin{center}
  \begin{tabular}{lcc}
    \textbf{} &{MNIST} & {SVHN} \\
    \midrule
$R$  & 0\% & 100\%   \\
$F$ & 100\% & 0\%  \\
  \bottomrule
\end{tabular}
\caption{Assignments of moderator to each of the competitors.}
\label{tab:mod_accuracy}
\end{center}
\end{table}

\subsubsection{Theoretical Analysis}
\label{appendix:convergence}

Competitive Collaboration is an alternating optimization procedure.
In the competition phase, we minimize $E_{1}$ with respect to $x$; in the collaboration phase we minimize $E_{2} = E_{1} + \lambda L_M$ with respect to $y$.
One might rightfully worry about the convergence properties of such a procedure, where we optimize different objectives in the alternating steps.

It is important to note that---while $E_{1}$ and $E_{2}$ are different functions---they are in fact closely related.
For example, they have the same minimizer with respect to the moderator output, namely assigning all the mass to the network with lower loss.
Ideally, we would want to use $E_{1}$ as the objective function in both phases, but resort to using $E_{2}$ in the collaboration phase, since it has empirically proven to be more efficient in pushing the moderator towards this optimal choice.

Hence, while we are minimizing different objective functions in the competition and collaboration phases, they are closely related and have the same ``goal''.
In the following, we formalize this mathematically by identifying general assumptions on how ``similar'' two functions have to be for such an alternating optimization procedure to converge.
Roughly speaking, we need the gradients of the two objectives to form an acute angle and to be of similar scales.
We will then discuss to what extent these assumptions are satisfied in the case of Competitive Collaboration.
Proofs are outsourced to the end of this section for readability.

\paragraph{General Convergence Theorem}

Assume we have two functions
\begin{equation}
f, g\colon \mathbb{R}^n \times \mathbb{R}^m \to \mathbb{R}
\end{equation}
and are performing alternating gradient descent updates of the form
\begin{gather}
x_{t+1} = x_t - \alpha \nabla_x f(x_t, y_t),\\
y_{t+1} = y_t - \beta \nabla_y g(x_{t+1}, y_t).
\end{gather}
We consider the case of single alternating gradient descent for convenience in the analysis.
With minor modifications, the following analysis also extends to the case of multiple gradient descent updates (or even exact minimization) in each of the alternating steps.
The following Theorem formulates assumptions on $f$ and $g$ under which such an alternating optimization procedure  converges to a first-order stationary point of $f$.

\begin{theorem}
\label{convergence_theorem}
Assume $f$ is lower-bounded and $x\mapsto \nabla_x f(x, y)$ is Lipschitz continuous with constant $G_1$ for every $y$ and $y\mapsto \nabla_y f(x, y)$ is Lipschitz continuous with constant $G_2(x)$.
Assume $\alpha \leq 2L_1^{-1}$.
If there is a constant $B>0$ such that
\begin{equation}
\label{eq:inner_product_assumption}
\begin{split}
\beta  \langle \nabla_y f(x, y), \nabla_y g(x, y) \rangle \geq & \frac{G_2(x) \beta^2}{2} \Vert \nabla_y g(x, y)\Vert^2 \\
& + B \Vert \nabla_y f(x, y)\Vert^2
\end{split}
\end{equation}
then $(x_t, y_t)$ converges to a first-order stationary point of $f$.
\end{theorem}

Eq.~\eqref{eq:inner_product_assumption} is a somewhat technical assumption that lower-bounds the inner product of the two gradients in terms of their norms and, thus, encodes that these gradients have to form an acute angle and be of similar scales.

\paragraph{Convergence of Competitive Collaboration}

We now discuss to what extent the assumptions for Theorem~\ref{convergence_theorem} are satisfied in the case of Competitive Collaboration.
For the mathematical considerations to follow, we introduce a slightly more abstract notation for the objective functions of Competitive Collaboration.
For a \emph{single data point}, $E_{1}$ has the form
\begin{equation}
\label{eq:abstract_competition_loss}
f(x, y) = M(y) L_R(x) + (1- M(y)) L_F(x),
\end{equation}
where $M(y)\in [0,1]$ is a function of $y$ (the weights of the moderator) and $L_R(x), L_F(x)>0$ are functions of $x$ (the weights of the two competing networks).
The loss function $E_{2}$ reads
\begin{equation}
\label{eq:abstract_collaboration_loss}
\begin{split}
& g(x, y) = f(x, y) \\
& + \lambda \cdot \begin{cases}
- \log( M(y) + \varepsilon) & \text{if } L_R(x) < L_F(x),\\
- \log(1-M(y) + \varepsilon) & \text{if } L_R(x) \geq L_F(x).
\end{cases}
\end{split}
\end{equation}
The following Proposition shows that $f$ and $g$ satisfy the conditions of Theorem~\ref{convergence_theorem} under certain assumptions.

\begin{proposition}
\label{proposition_single_data_point_cc}
Let $f$ and $g$ be defined by Equations \eqref{eq:abstract_competition_loss} and \eqref{eq:abstract_collaboration_loss}, respectively.
If $M(y)$, $L_R(x)$ and $L_F(x)$ are Lipschitz smooth, then $f$ and $g$ fulfill the assumptions of Theorem 1.
\end{proposition}

The smoothness conditions on $M(y)$, $L_R(x)$, $L_F(x)$ are standard as they are, for example, needed to guarantee convergence of gradient descent for optimizing any of these objective functions individually.

This Proposition shows that the objectives for individual data points satisfy Theorem 1.
In practice, however, we are concerned with multiple data points and objectives of the form
\begin{equation}
f(x, y) = \frac{1}{n} \sum_{i=1}^n f^{(i)}(x, y),
\end{equation}
where
\begin{equation}
\begin{split}
f^{(i)}(x, y) = & M^{(i)}(y) L_R^{(i)}(x)\\
&+ (1- M^{(i)}(y)) L_F^{(i)}(x),
\end{split}
\end{equation}
and
\begin{equation}
g(x, y) = \frac{1}{n} \sum_{i=1}^n g^{(i)}(x, y),
\end{equation}
where
\begin{equation}
\begin{split}
& g^{(i)}(x, y) = f^{(i)}(x, y) \\
& + \lambda \cdot \begin{cases}
- \log( M^{(i)}(y) + \varepsilon) & \text{if } L_R^{(i)}(x) < L_F^{(i)}(x),\\
- \log(1-M^{(i)}(y) + \varepsilon) & \text{if } L_R^{(i)}(x) \geq L_F^{(i)}(x).
\end{cases}
\end{split}
\end{equation}
While we have just found a suitable lower bound on the inner product of $\nabla_y f^{(i)}$ and $\nabla_y g^{(i)}$, unfortunately, the sum structure of $\nabla_y f$ and $\nabla_y g$ makes it really hard to say anything definitive about the value of their inner product.
It is \emph{plausible} to assume that $\nabla_y f$ and $\nabla_y g$ will be sufficiently close to guarantee convergence in practical settings.
However, the theory developed in Theorem \ref{convergence_theorem} does not directly apply.

\subsubsection{Proofs}

\label{proofs}

\begin{proof}[\textbf{Proof of Theorem}~\ref{convergence_theorem}]
The update of $x$ is a straight-forward gradient descent step on $f$.
Using the Lipschitz bound on $f$, we get
\begin{equation}
\begin{split}
f(x_{t+1}, y_t) & \leq f(x_t, y_t) -\alpha \langle \nabla_x f(x_t, y_t), \nabla_x f(x_t, y_t)\rangle \\
& + \frac{G_1 \alpha^2}{2} \Vert \nabla_x f(x_t, y_t)\Vert^2 \\
& = f(x_t, y_t) - \left( \alpha - \frac{G_1\alpha^2}{2} \right) \Vert \nabla_x f(x_t, y_t) \Vert^2\\
& \leq f(x_t, y_t) - A \Vert\nabla_x f(x_t, y_t)\Vert^2
\end{split}
\end{equation}
with $A>0$ due to our assumption on $\alpha$.
For the update of $y$, we have
\begin{equation}
\begin{split}
f(x_{t+1}, y_{t+1}) &\leq f(x_{t+1}, y_t) \\
& - \beta \langle \nabla_y f(x_{t+1}, y_t), \nabla_y g(x_{t+1}, y_t) \rangle\\
& + \frac{\beta^2 G_2(x)}{2} \Vert \nabla_y g(x_{t+1}, y_t)\Vert^2.
\end{split}
\end{equation}
Using the assumption on the inner product, this yields
\begin{equation}
\begin{split}
f(x_{t+1}, y_{t+1}) & \leq f(x_{t+1}, y_t) - B \Vert \nabla_y f(x_{t+1}, y_t) \Vert^2.
\end{split}
\end{equation}
Combining the two equations, we get
\begin{equation}
\label{eq:combined_improvement}
\begin{split}
& f(x_{t+1}, y_{t+1}) \\
& \leq f(x_t, y_t) - A \Vert \nabla_x f(x_t, y_t) \Vert^2  - B \Vert \nabla_y f(x_{t+1}, y_t) \Vert^2\\
& \leq f(x_t, y_t) - C \left( \Vert \nabla_x f(x_t, y_t) \Vert^2 + \Vert \nabla_y f(x_{t+1}, y_t) \Vert^2 \right).
\end{split}
\end{equation}
with $C=\max (A, B)$.
We define $G_t= \Vert \nabla_x f(x_t, y_t) \Vert^2 + \Vert \nabla_y f(x_{t+1}, y_t) \Vert^2$ and rewrite this as
\begin{equation}
G_t \leq \frac{f(x_t, y_t) - f(x_{t+1}, y_{t+1})}{C}
\end{equation}
Summing this equation for $t=0,\dotsc, T$, we get
\begin{equation}
\sum_{t=0}^T G_t \leq \frac{f(x_0, y_0) - f(x_{T+1}, y_{T+1})}{C}.
\end{equation}
Since $f$ is lower-bounded, this implies $G_t \rightarrow 0$, which in turn implies convergence to a first-order stationary point of $f$.
\end{proof}

\begin{proof}[\textbf{Proof of Proposition} \ref{proposition_single_data_point_cc}]
The gradient of $f$ with respect to $x$ is
\begin{equation}
\nabla_x f(x, y) = M(y) \nabla L_R(x) + (1-M(y)) \nabla L_F(x)
\end{equation}
Since $M(y)$ is bounded, $\nabla_x f$ is Lipschitz continuous in $x$ given that $L_R$ and $L_F$ are Lipschitz smooth.

For the assumptions on the $y$-gradients, we fix $x$ and treat the two cases in the definition of $g$ separately.
We only consider the case $L_R(x)< L_F(x)$ here, the reverse case is completely analogous.
Define $L(x) = L_F(x) - L_R(x) >0$.
The gradient of $f$ with respect to $y$ is
\begin{equation}
\nabla_y f(x, y) = -L(x) \nabla M(y)
\end{equation}
and is Lipschitz continuous with constant $G_2(x) = \vert -L(x)\vert G = L(x)G$, where $G$ is the Lipschitz constant of $M(y)$.
We have
\begin{equation}
\nabla_y g(x, y) = - \left(L(x) + \frac{\lambda}{M(y) + \varepsilon} \right) \nabla M(y).
\end{equation}
The inner product of the two gradients reads
\begin{equation}
\begin{split}
& \langle \nabla_y f(x, y), \nabla_y g(x, y)\rangle \\
& = L(x) \left( L(x) + \frac{\lambda}{M(y) + \varepsilon} \right) \Vert \nabla M(y) \Vert^2,
\end{split}
\end{equation}
and for the gradient norms we get
\begin{equation}
\Vert \nabla_y f(x, y)\Vert^2 = L(x)^2 \Vert \nabla M(y)\Vert^2,
\end{equation}
as well as
\begin{equation}
\Vert \nabla_y g(x, y)\Vert^2 = \left( L(x) + \frac{\lambda}{M(y)+ \varepsilon} \right)^2 \Vert \nabla M(y)\Vert^2.
\end{equation}
Plugging everything into the inner product assumption of Theorem 1 and simplifying yields
\begin{equation}
\begin{split}
\beta \left( L(x) + \frac{\lambda}{M(y)+\varepsilon} \right) & \geq \frac{G\beta^2}{2} \left( L(x) + \frac{\lambda}{M(y)+\varepsilon} \right)^2 \\
& + BL(x)
\end{split}
\end{equation}
Since $M$, $L_R$ and $L_F$ are bounded, one easily finds a choice for $\beta$ and $B$ that satisfies this condition.
\end{proof}

\subsection{The camera warping function $w_c$ and static flow transformer $\nu$}
\label{appendix:camera}
The network $C$ predicts camera motion that consist of camera rotations $sin\alpha, sin \beta, sin \gamma $, and translations $t_x, t_y, t_z $. 
Thus $e = (sin\alpha, sin \beta, sin \gamma, t_x, t_y, t_z)$. 
Given camera motion and depth $d$, we transform the image coordinates $(x,y)$ into world coordinates $(X,Y,Z)$.
\begin{eqnarray}
&X = \frac{d}{f}(x - c_x) \\
&Y = \frac{d}{f}(y - c_y) \\
&Z = d
\end{eqnarray}
where $(c_x, c_y, f)$ constitute the camera intrinsics.
We now transform the world coordinates given the camera rotation and translation.
$$\mathbf{X'} = R_xR_yR_z\mathbf{X} + t$$
where $(R_xR_yR_z, t) \in SE3$ denote 3D rotation and translation,
and $\mathbf{X} = [X,Y,Z]^T$.
Hence, in image coordinates
\begin{eqnarray}
x' = \frac{f}{Z} + c_x \\
y' = \frac{f}{Z} + c_y
\end{eqnarray}
We can now apply the warping as,
\begin{equation}
w_c \big(I(x,y),e,d\big) = I(x',y').
\end{equation}
The static flow transformer is defined as,
\begin{equation}
\nu(e,d) = (x'-x, y'-y)
\end{equation}

\subsection{The flow warping function, $w_f$}
\label{appendix:flow}
The flow warping function $w_f$ is given by
\begin{equation}
w_f\big(I(x,y), u_x,u_y\big) = I(x+u_x, y+u_y)
\end{equation}
where, $(u_x,u_y) $ is the optical flow, and $(x,y)$ is the spatial coordinate system.

\subsection{Network Architectures}
\label{appendix:nets}
\begin{figure*}[t]
  \centering
	\includegraphics[width=0.84\linewidth]{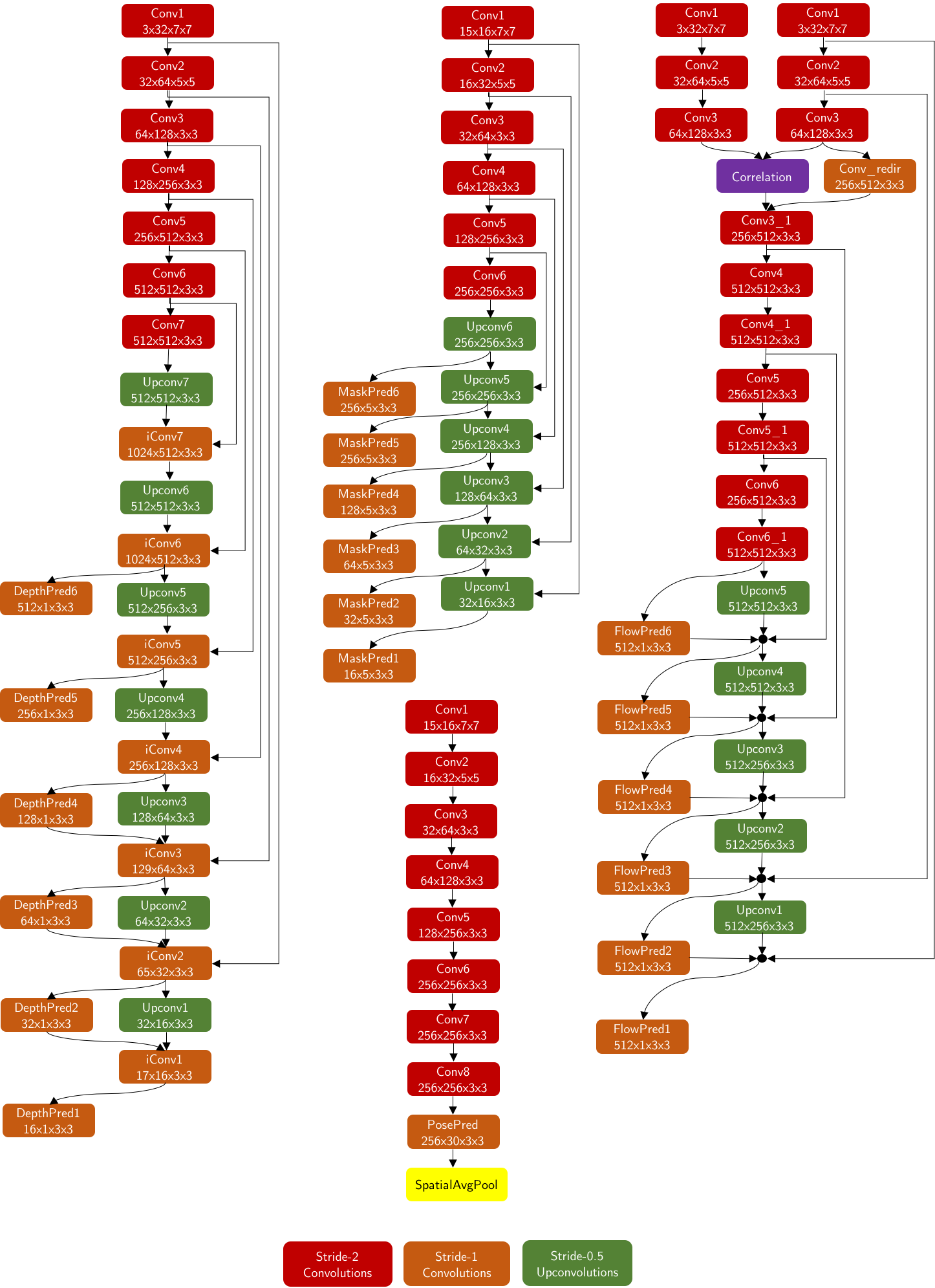}
	\caption{Architecture of the DispNet (left), MaskNet (center-top), FlowNetC (right) and Camera Motion Network (center-bottom). Convolutional layers are red (stride 2) and orange (stride 1) and upconvolution layers are green (stride 2). Other colors refer to special layers. Each layer is followed by ReLU, except prediction layers. In each block, the numbers indicate the number of channels of the input feature map, the number of channels of the output feature map, and the filter size.}
    \label{fig:nets}
\end{figure*}

We briefly describe the network architectures below. For details, please refer to Figure \ref{fig:nets}.

\textbf{Depth Network $D$.} Our depth network is similar to DispNetS \cite{mayer2016large} and outputs depths at 6 different scales. Each convolution and upconvolution is followed by a ReLU except the prediction layers. The prediction layer at each scale has a non-linearity given by $1/(\alpha$ sigmoid$(x) + \beta) $. The architecture of DispResNet is obtained by replacing convolutional blocks in DispNet by residual blocks \cite{he2016resnet}.

\textbf{Camera Motion Network $C$.} The camera motion network consists of 8 convolutional layers, each of stride 2 followed by a ReLU activation. This is followed by a convolutional layer of stride 1, whose feature maps are averaged together to get the camera motion.

\textbf{Flow Network $F$.} We use FlowNetC architecture \cite{dosovitskiy2015flownet} with 6 output scales of flow and is shown in Figure \ref{fig:nets}. All convolutional and upconvolutional layers are followed by a ReLU except prediction layers. The prediction layers have no activations. For PWC Net, we use the network architecture from Janai et al. \cite{janai2018unsupervised}.

\textbf{Mask Network $M$.}
The mask network has a U-Net \cite{dosovitskiy2015flownet} architecture. The encoder is similar to the camera motion with 6 convolutional layers. The decoder has 6 upconvolutional layers. Each of these layers have ReLU activations. The prediction layers use a sigmoid.

\begin{figure}[t]
\begin{center}
   \includegraphics[width=\linewidth]{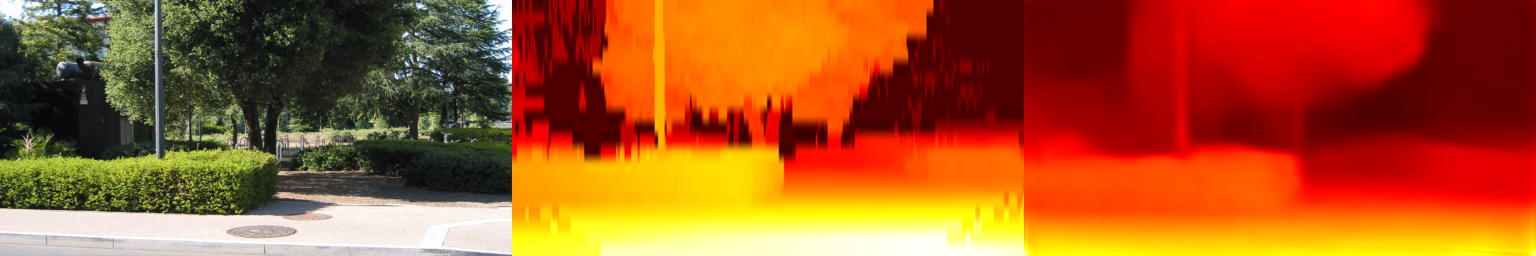}
   \includegraphics[width=\linewidth]{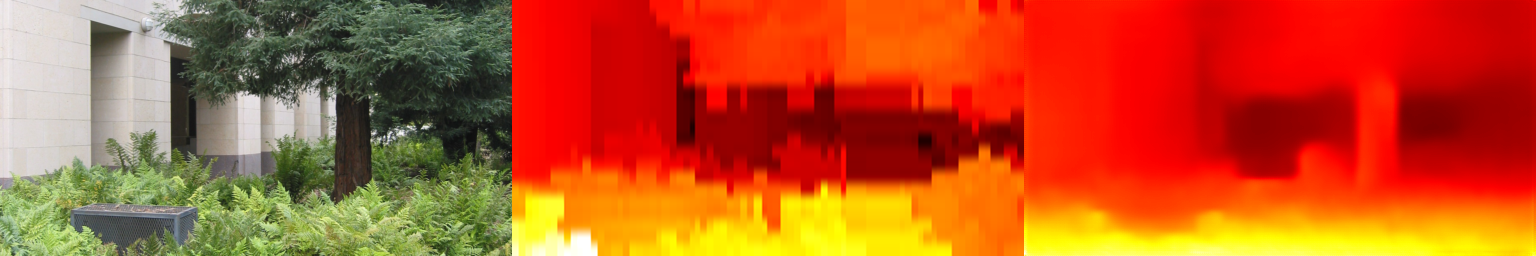}
   \includegraphics[width=\linewidth]{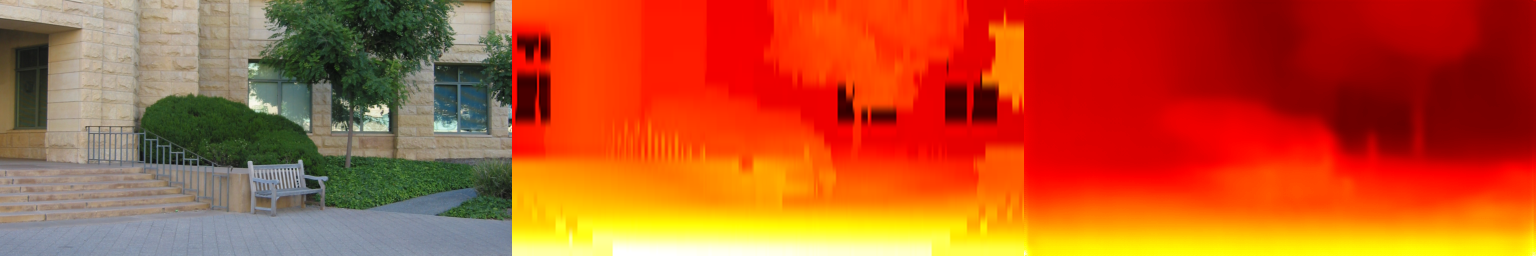}
   \includegraphics[width=\linewidth]{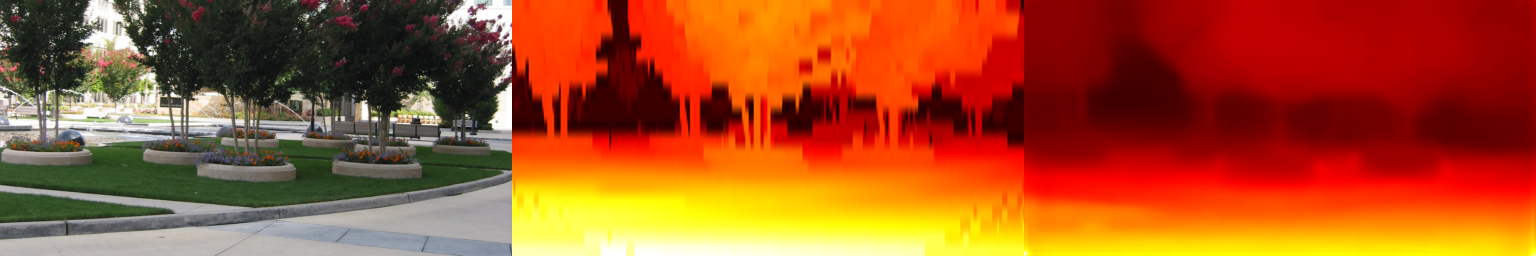} \\
   \small{\quad Image \quad \quad \quad \quad \quad Ground Truth \quad \quad \quad \quad Prediction}
\end{center}
   \caption{Qualitative results on Make3D test set.}
\label{fig:make3d}
\end{figure}

\begin{figure*}[t]
  \centering
 \begin{tabular}{C{0.3\textwidth}C{0.3\textwidth}C{0.3\textwidth}}
Image  & Predicted Depth& Consensus Mask \\
\end{tabular}
\includegraphics[width=0.9\linewidth]{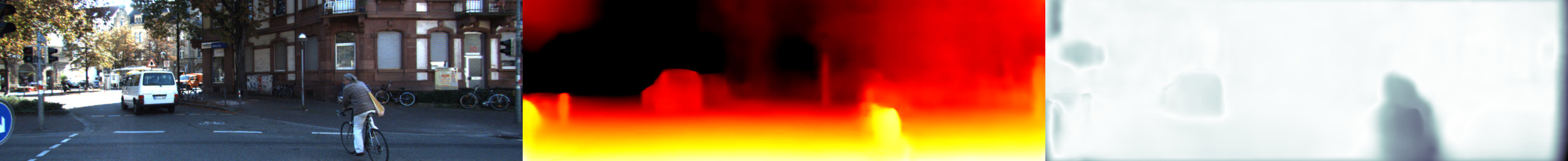} \\
\begin{tabular}{C{0.3\textwidth}C{0.3\textwidth}C{0.3\textwidth}}
Static scene Optical Flow  & Segmented Flow in moving regions& Full Optical Flow \\
\end{tabular}
\includegraphics[width=0.9\linewidth]{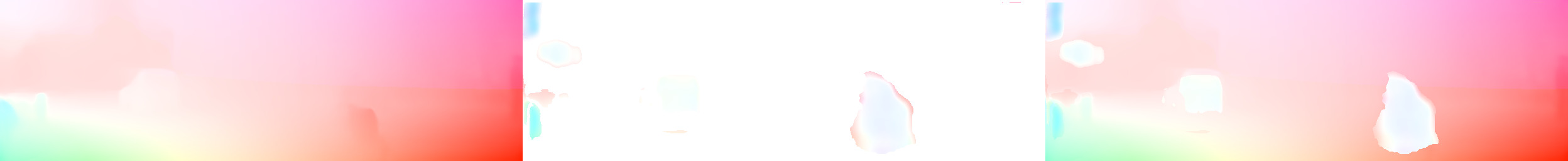} \\ \vspace*{0.06in}
\includegraphics[width=0.9\linewidth]{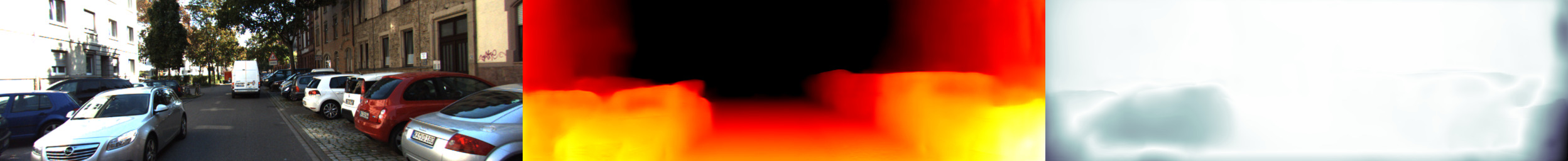} \\
\includegraphics[width=0.9\linewidth]{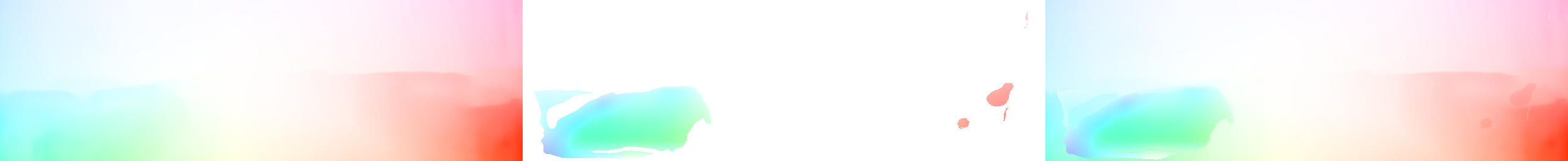} \\ \vspace*{0.06in}
\includegraphics[width=0.9\linewidth]{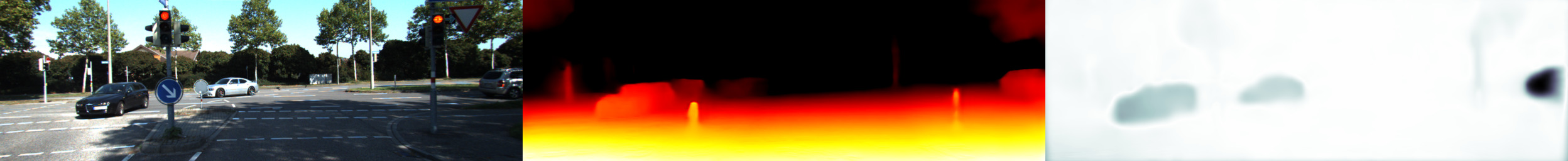} \\
\includegraphics[width=0.9\linewidth]{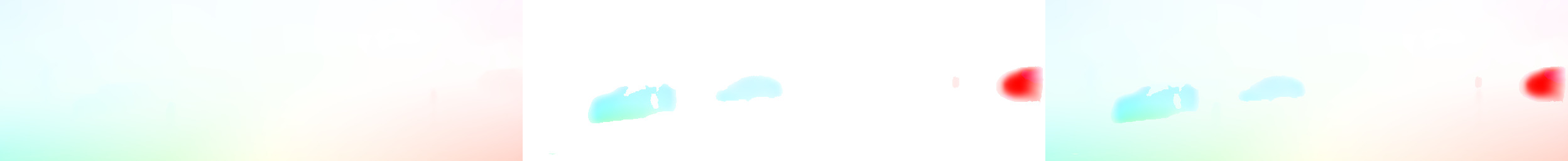} \\ \vspace*{0.06in}
\includegraphics[width=0.9\linewidth]{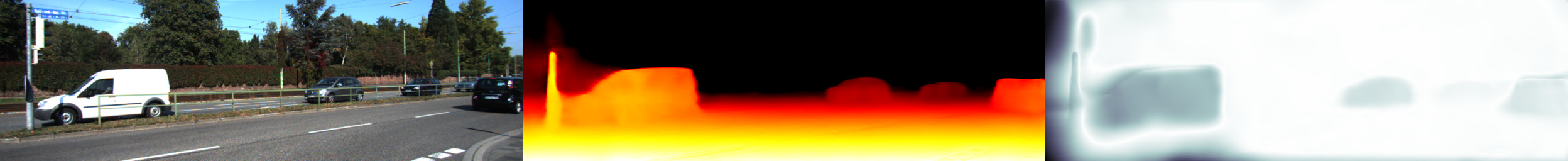} \\
\includegraphics[width=0.9\linewidth]{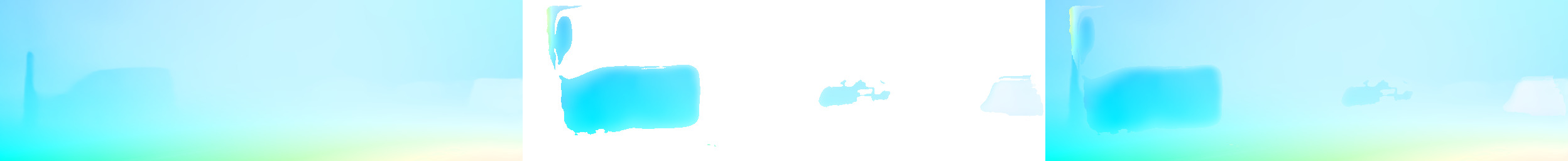} \\ \vspace*{0.06in}
\includegraphics[width=0.9\linewidth]{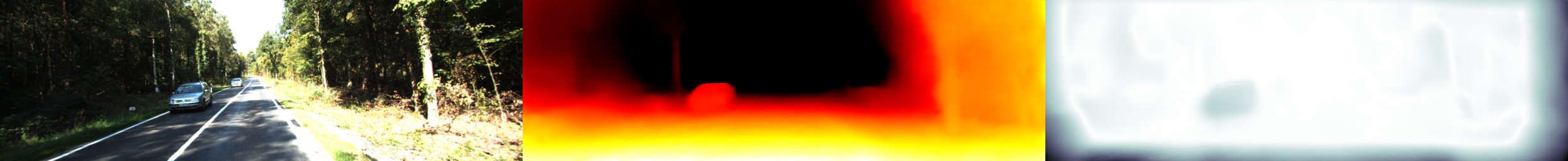} \\
\includegraphics[width=0.9\linewidth]{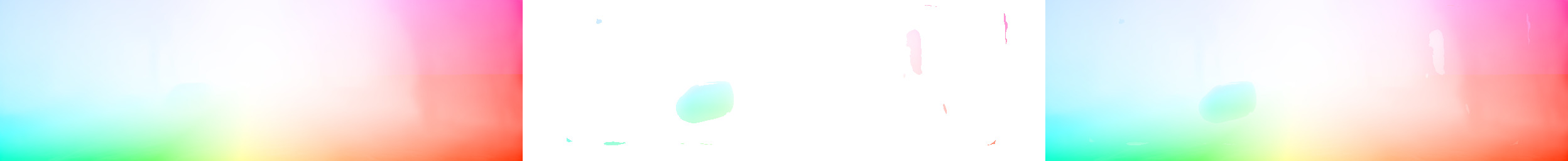} \\	\vspace*{0.06in}
\includegraphics[width=0.9\linewidth]{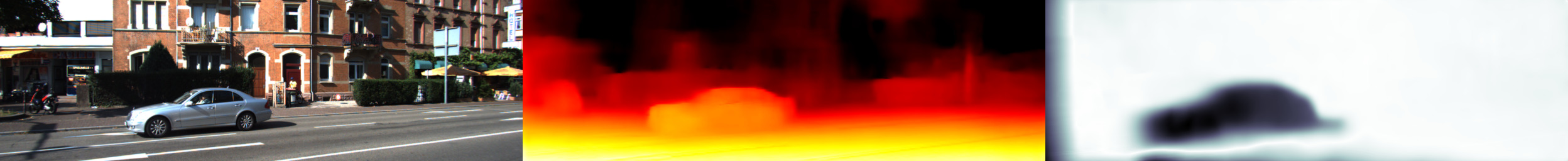} \\
\includegraphics[width=0.9\linewidth]{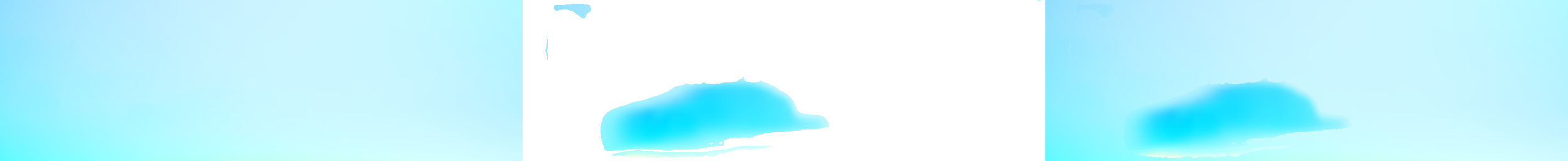} \\ \vspace*{0.06in}
	\caption{Network Predictions. Top row: we show image, predicted depth, consensus masks. Bottom row: we show static scene optical flow, segmented flow in the moving regions and full optical flow. }
    \label{fig:viz_extra}
\end{figure*}

\begin{figure*}[t]
  \centering
 \begin{tabular}{C{0.3\textwidth}C{0.3\textwidth}C{0.3\textwidth}}
Image  & Ground Truth & Zhou et al. \cite{zhou2017unsupervised} \\
\end{tabular}
\includegraphics[width=0.9\linewidth, trim={0 5cm 0 0},clip]{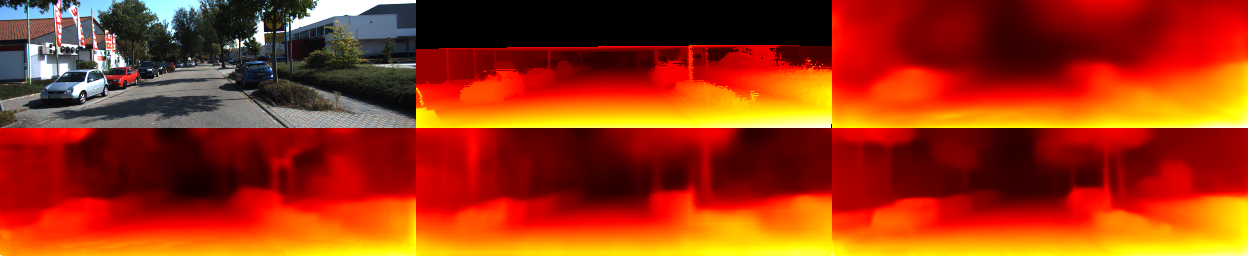} \\
\begin{tabular}{C{0.3\textwidth}C{0.3\textwidth}C{0.3\textwidth}}
Geonet \cite{yin2018geonet}  & DF-Net \cite{zou2018df} & CC (ours)\\
\end{tabular}
\includegraphics[width=0.9\linewidth, trim={0 0 0 5cm},clip]{depth_compare_0027.png} \\ \vspace*{0.06in}
\includegraphics[width=0.9\linewidth]{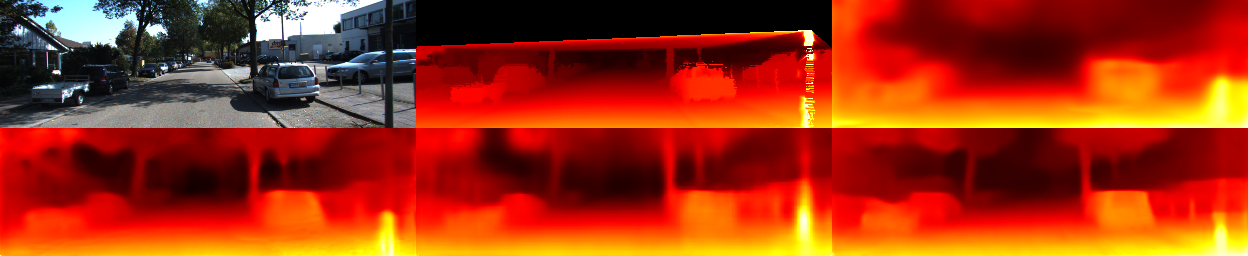} \\ \vspace*{0.06in}
\includegraphics[width=0.9\linewidth]{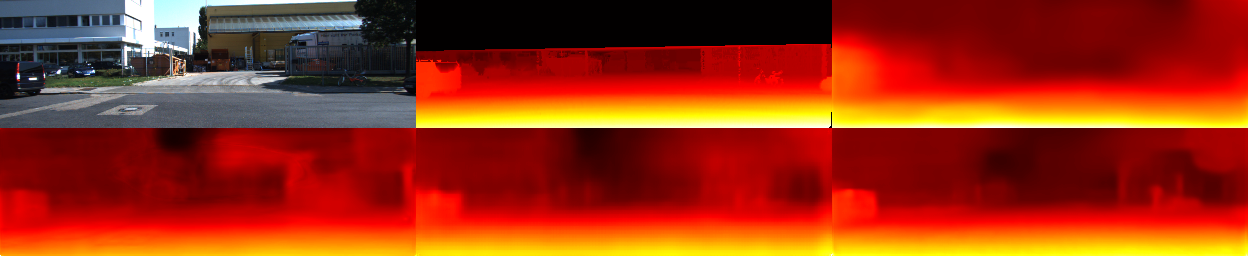} \\ \vspace*{0.06in}
\includegraphics[width=0.9\linewidth]{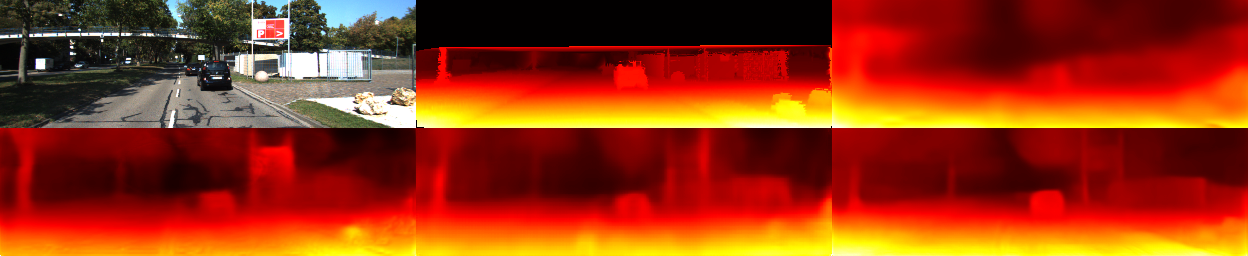} \\ \vspace*{0.06in}
\includegraphics[width=0.9\linewidth]{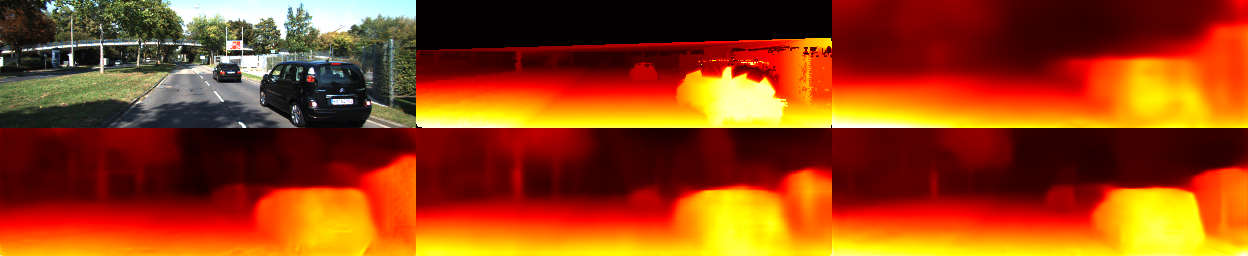} \\	\vspace*{0.06in}
\includegraphics[width=0.9\linewidth]{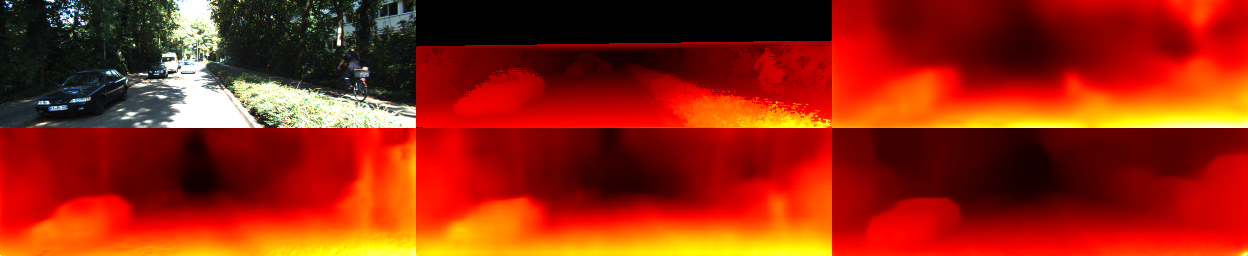} \\ \vspace*{0.06in}

	\caption{Qualitative results on single view depth prediction. Top row: we show image, interpolated ground truth depths, Zhou et al. \cite{zhou2017unsupervised} results. Bottom row: we show results from Geonet \cite{yin2018geonet}, DF-Net \cite{zou2018df} and CC (ours) results. }
    \label{fig:depth_extra}
\end{figure*}

\begin{figure*}[t]
  \centering
 \begin{tabular}{C{0.3\textwidth}C{0.3\textwidth}C{0.3\textwidth}}
Image  & Ground Truth & CC (DispResNet)  \\
\end{tabular}
\includegraphics[width=0.9\linewidth, trim={0 5cm 0 0},clip]{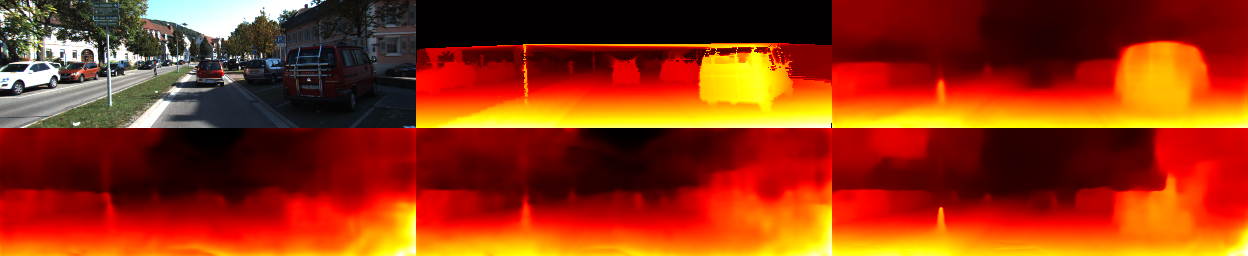} \\
\begin{tabular}{C{0.3\textwidth}C{0.3\textwidth}C{0.3\textwidth}}
Basic (DispNet)  & Basic+ssim (DispNet) & CC (DispNet)\\
\end{tabular}
\includegraphics[width=0.9\linewidth, trim={0 0 0 5cm},clip]{depth_ablation_0311.png} \\ \vspace*{0.06in}
\includegraphics[width=0.9\linewidth]{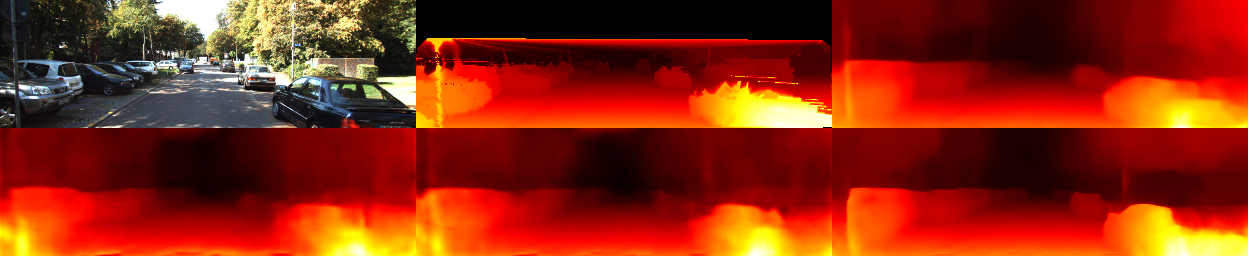} \\ \vspace*{0.06in}
\includegraphics[width=0.9\linewidth]{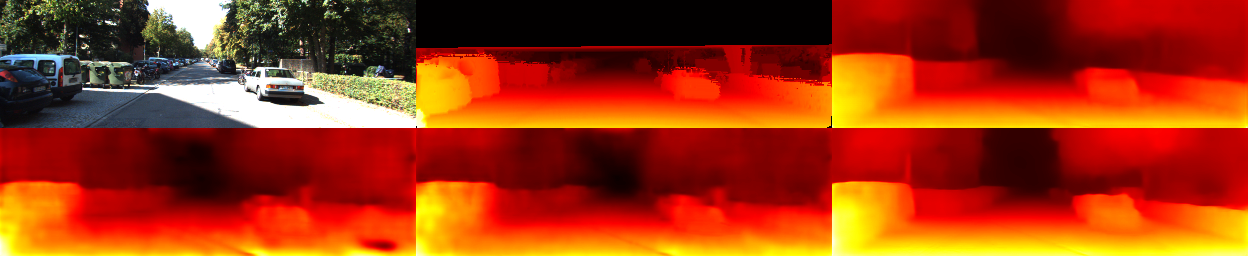} \\ \vspace*{0.06in}
\includegraphics[width=0.9\linewidth]{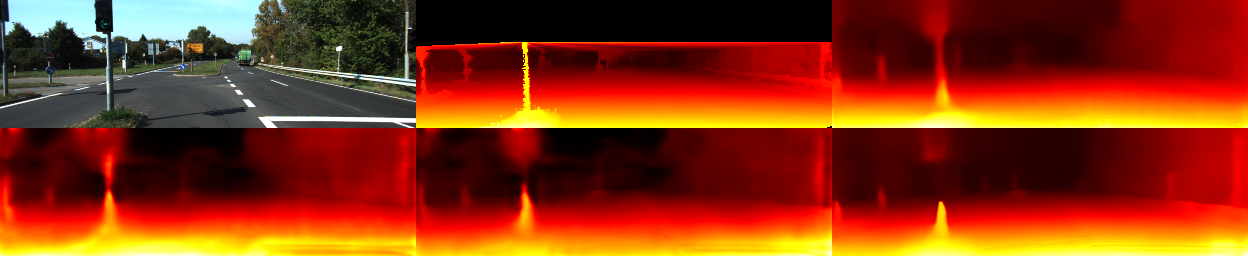} \\ \vspace*{0.06in}
\includegraphics[width=0.9\linewidth]{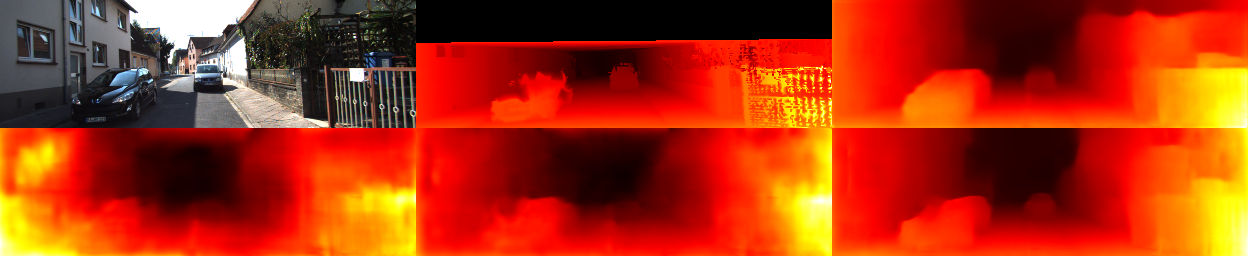} \\	\vspace*{0.06in}
\includegraphics[width=0.9\linewidth]{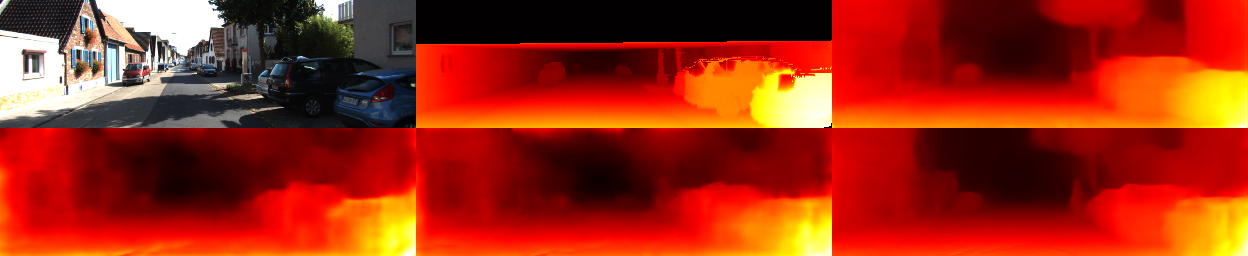} \\ \vspace*{0.06in}

	\caption{Ablation studies on single view depth prediction. Top row: we show image, interpolated ground truth depths, CC using DispResNet architecture. Bottom row: we show results using Basic, Basic+ssim and CC models using DispNet architecture. }
    \label{fig:depth_ablation}
\end{figure*}

\begin{figure*}[t]
  \centering
 \begin{tabular}{C{0.3\textwidth}C{0.3\textwidth}C{0.3\textwidth}}
Image  & Ground Truth & UnFlow-CSS \cite{meister2017unflow} \\
\end{tabular}
\includegraphics[width=0.9\linewidth, trim={0 5cm 0 0},clip]{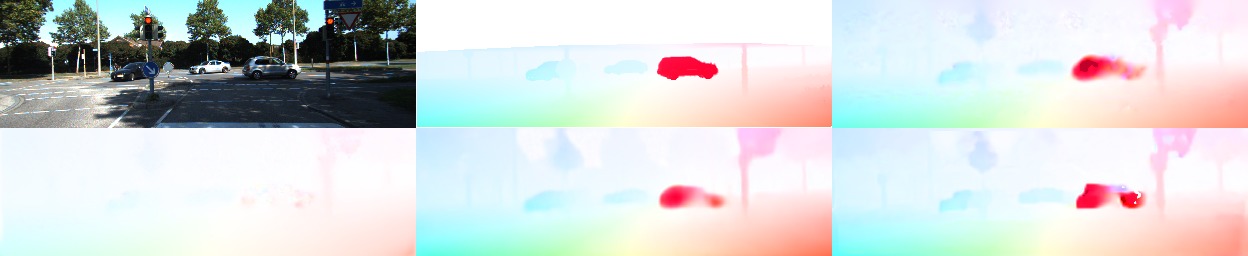} \\
\begin{tabular}{C{0.3\textwidth}C{0.3\textwidth}C{0.3\textwidth}}
Geonet \cite{yin2018geonet}  & DF-Net \cite{zou2018df} & CC (ours)\\
\end{tabular}
\includegraphics[width=0.9\linewidth, trim={0 0 0 5cm},clip]{flow_0010.jpg} \\ \vspace*{0.06in}
\includegraphics[width=0.9\linewidth]{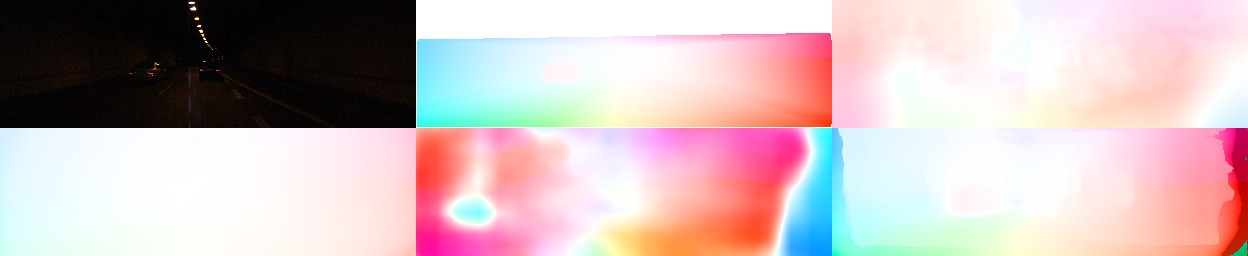} \\ \vspace*{0.06in}
\includegraphics[width=0.9\linewidth]{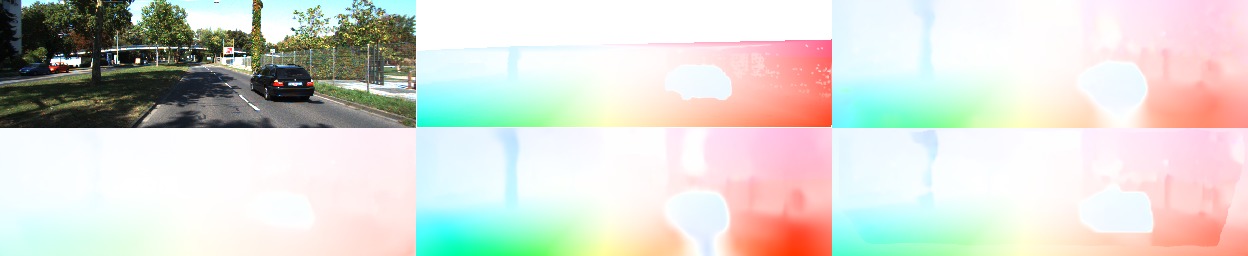} \\ \vspace*{0.06in}
\includegraphics[width=0.9\linewidth]{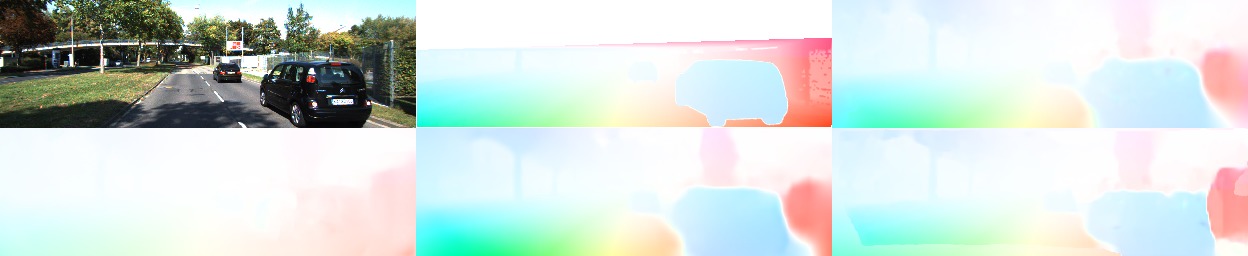} \\ \vspace*{0.06in}
\includegraphics[width=0.9\linewidth]{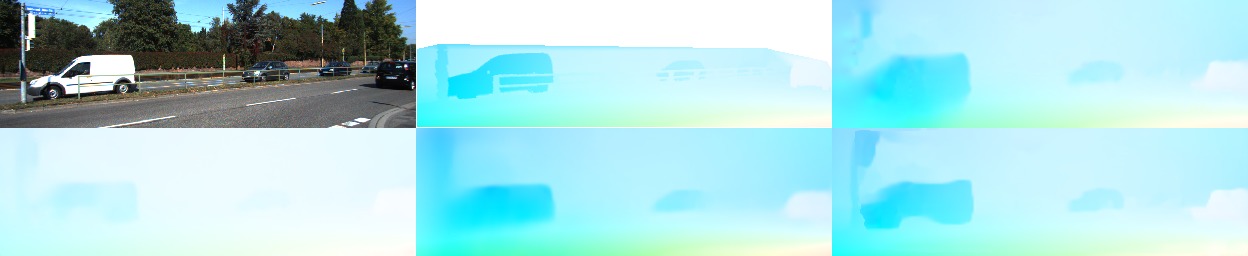} \\	\vspace*{0.06in}
\includegraphics[width=0.9\linewidth]{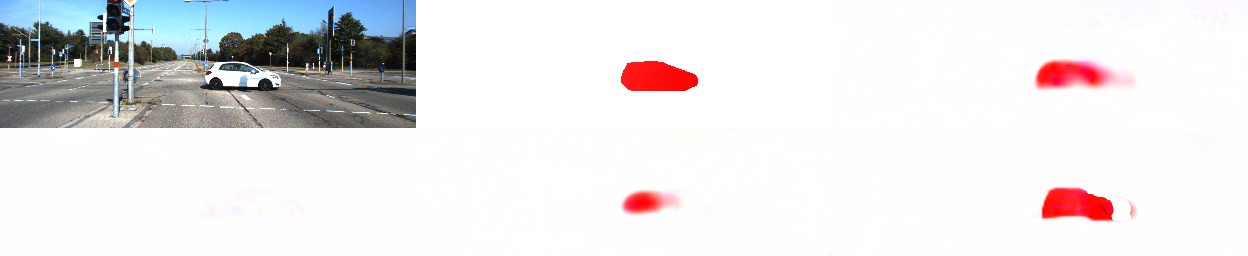} \\ \vspace*{0.06in}

	\caption{Qualitative results on Optical Flow estimation. Top row: we show image 1, ground truth flow, and predictions from  UnFlow-CSS\cite{meister2017unflow}. Bottom row: we show predictions from Geonet \cite{yin2018geonet}, DF-Net \cite{zou2018df} and CC (ours) model. }
    \label{fig:flow_extra}
\end{figure*}

\subsection{Qualitative Results}
\label{appendix:results}
The qualitative results of the predictions are shown in Figure \ref{fig:viz_extra}. We would like to point out that, our method is able to segment the moving car, and not the parked cars on the roads. In addition, it segments other moving objects, such as the bicyclist.

We compare the qualitative results for single image depth prediction in Figure \ref{fig:depth_extra}. We also contrast our results with basic models that were trained independently without a joint loss in Figure \ref{fig:depth_ablation}. We observe that our model produces better results, capturing moving objects such as cars and bikes, as well as surface edges of trees, pavements and buildings.

We compare the qualitative results for optical flow estimation in Figure \ref{fig:flow_extra}. We show that our method performs better than UnFlow \cite{meister2017unflow}, Geonet \cite{yin2018geonet} and DF-Net \cite{zou2018df} . Our flow estimations are better at the boundaries of the cars and pavements. In contrast, competing methods produce blurry flow fields.

\subsection{Additional Experiments}
\label{appendix:extra}
\paragraph{Depth evaluation on Make3D dataset.}
We also test on the Make3D dataset \cite{saxena2006learning} without training on it. We use our our model that is trained only on Cityscapes and KITTI. Our method outperforms previous work \cite{zhou2017unsupervised, zou2018df, godard2018digging} as shown in Table~\ref{tab:depth_make3d}. We show qualitative results in Fig.~\ref{fig:make3d}.

\paragraph{Pose evaluation on Sintel.}
We test on Sintel's alley sequence \cite{butler2012naturalistic} without training on it and compare it with Zhou et al.~\cite{zhou2017unsupervised}. For this comparison, Zhou et al.'s model is taken from Pinard's implementation. We show quantitative evaluation using relative errors on pose in Table \ref{tab:sintel_pose}.

\paragraph{Training using a shared encoder.}
We train the camera motion network, $C$ and motion segmentation network, $M$ using a common shared encoder but different decoders. Intuitively, it seems that camera motion network can benefit from knowing static regions in a scene, which are learned by the motion segmentation network. However, we observe a performance degradation on camera motion estimates (Table \ref{tab:pose_shared}).  The degradation of results using a shared encoder are because feature encodings for one network might not be optimal for other networks.
Our observation is consistent with Godard et al. \cite{godard2018digging} (Supp. Mat. Table 4), where sharing an encoder for depth and camera motion estimation improves depth but the perfomance on camera motion estimates are not as good.

\begin{table}
\begin{center}
\begin{tabular}{cccc}
 {Zhou} \cite{zhou2017unsupervised} & {DF-Net} \cite{zou2018df} & {Godard} \cite{godard2018digging} &  {CC (ours)} \\ \hline
0.383 & 0.331 &  0.361 & \textbf{0.320}
\end{tabular}
\caption{Absolute Relative errors on Make3D test set.}
\label{tab:depth_make3d}
\end{center}
\end{table}

\begin{table}
\begin{center}
\begin{tabular}{lcc}
& alley 1 & alley 2 \\ \hline
 {Zhou et al.} \cite{zhou2017unsupervised} & $ 0.002  \pm 0.001$ & $0.027 \pm 0.019 $ \\
CC (ours) & $ {0.002}  \pm {0.001}$ &  ${0.022} \pm {0.015} $
\end{tabular}
\caption{Relative errors on Sintel alley sequences.}
\label{tab:sintel_pose}
\end{center}
\end{table}

\begin{table}
\begin{center}
\begin{tabular}{lcc}
  & Sequence 09 & Sequence 10 \\ \hline
 {Shared Encoder} & 0.017 $\pm$ 0.009  & 0.015 $\pm$ 0.009\\
{Uncoupled Networks} & \textbf{0.012} $\pm$ \textbf{0.007} & \textbf{0.012} $\pm$ \textbf{0.008} \\
\end{tabular}
\caption{Absolute Trajectory errors on KITTI Odometry.}
\label{tab:pose_shared}
\end{center}
\end{table}

\subsection{Timing Analysis}
We analyze inference time of our network and compare it with  Geonet~\cite{yin2018geonet} in Table \ref{tab:inference}. We observe that our networks have a faster run time using the same sized 128 $\times$416 images on a single TitanX GPU. This is because our networks are simpler and smaller than ones used by Geonet.

For training, we measure the time taken for each iteration consisting of forward and backward pass using a batch size of 4. Training depth and camera motion networks $(D,C)$ takes $0.96s$ per iteration. Traing the mask network, $M$ takes $0.48s$ per iteration, and the flow network $F$ takes $1.32s$ per iteration. All iterations have a  batch size of 4. In total, it takes about 7 days for all the networks to train starting with random initialization on a single 16GB Tesla V100.

\begin{table}
\begin{center}
\begin{tabular}{lcccc}
\small{Method} & \small{Depth} & \small{Pose} & \small{Flow} & \small{Mask}  \\ \hline
\small{Geonet} \cite{yin2018geonet}  & $15ms$ & $4ms$& $45ms$ & -\\
CC (ours) & $13ms$ & $2ms$ & $34ms$ & $3ms$\\
\end{tabular}
\caption{Average runtime on TitanX GPU with images of size 128 $\times$ 418.}
\label{tab:inference}
\end{center}
\end{table}

\end{appendices}

\end{document}